%% file: acl_latex.tex
\pdfoutput=1

\documentclass[11pt]{article}

\usepackage{acl}

\usepackage{times}
\usepackage{latexsym}
\usepackage{graphicx}
\usepackage{xspace}
\usepackage{mathtools}
\usepackage{multirow}
\usepackage{subcaption}

\usepackage[T1]{fontenc}

\usepackage[utf8]{inputenc}

\usepackage{microtype}

\usepackage[colorinlistoftodos]{todonotes}

\newcommand{\lambada}{LAMBADA\xspace}
\newcommand{\lpp}{LPP\xspace}

\newenvironment{enumeratesquish}{\begin{list}{\addtocounter{enumi}{1}\labelenumi}{\setlength{\itemsep}{0em}\setlength{\labelwidth}{0.5em}\setlength{\leftmargin}{\labelwidth}\addtolength{\leftmargin}{\labelsep}}}{\end{list}\setcounter{enumi}{0}}

\title{Improving In-Context Few-Shot Learning\\for Pretrained Language Models via Self-Supervised Training}
\title{An Empirical Study on Improving In-Context Few-Shot Learning\\via Self-Supervised Training}
\title{Improving In-Context Few-Shot Learning\\via Self-Supervised Training}

\author{
Mingda Chen$^1$\thanks{~~Work done during an internship at Meta AI.}\quad Jingfei Du$^2$\quad Ramakanth Pasunuru$^2$\quad Todor Mihaylov$^2$\quad\\ \textbf{Srini Iyer}$^2$\quad \textbf{Veselin Stoyanov}$^2$\quad \textbf{Zornitsa Kozareva}$^2$ \\
$^{1}$Toyota Technological Institute at Chicago, IL, USA\\
$^{2}$Meta AI \\
  \texttt{mchen@ttic.edu}\\
  \texttt{\{jingfeidu,rpasunuru,tbmihaylov,sviyer,ves,zori\}@fb.com}}

\begin{document}
\maketitle
\begin{abstract}
Self-supervised pretraining has made few-shot learning possible for many NLP tasks. But the pretraining objectives are not typically adapted specifically for in-context few-shot learning.
In this paper, we propose to use self-supervision in an intermediate training stage between pretraining and downstream few-shot usage with the goal to teach the model to perform in-context few shot learning. We propose and evaluate four self-supervised objectives on two benchmarks. We find that the intermediate self-supervision stage produces models that outperform strong baselines. Ablation study shows that several factors affect the downstream performance, such as the amount of training data and the diversity of the self-supervised objectives. Human-annotated cross-task supervision and self-supervision are complementary. Qualitative analysis suggests that the self-supervised-trained models are better at following task requirements.

\end{abstract}

\input{intro}
\input{related}

\input{method}
\input{experiment}

\input{analysis}
\input{conclusion}

\bibliography{anthology,custom}
\bibliographystyle{acl_natbib}
\clearpage

\appendix
\input{appendix}

\end{document}

%% file: intro.tex
\section{Introduction}
In-context few-shot learning seeks to solve unseen tasks at inference time by conditioning on a few training examples. In particular, in this case we are interested in methods that forgo any weight updates \citep{gpt3}. Prior work has been focused on improving inference time algorithms (e.g., rescoring generated outputs \citep{pmlr-v139-zhao21c}, selecting \citep{liu2021makes} and ordering \citep{lu2021fantastically} the given few-shot examples) and incorporating extra resources (e.g., fine-tuning models on human-annotated datasets \citep{mishra2021crosstask,ye2021crossfit,wei2021finetuned}). 

We hypothesise that a different way to improve in-context few-shot learning is through designing self-supervised objectives that more closely resemble the format of tasks that the model will be asked to perform. To do so, we cast the self-supervised training as an intermediate training stage between language model pretraining and downstream few-shot evaluation. In particular, we construct training datasets based on the self-supervised objectives following similar formats used in the downstream tasks, fine-tune pretrained language model checkpoints on the training datasets, and then evaluate the models on benchmarks.

In experiments, we consider four self-supervised objectives, including masked word prediction and classification tasks related to next sentence prediction \citep{devlin-etal-2019-bert}. We evaluate models on two benchmarks (13 tasks in total): SuperGLUE \citep{superglue} and Natural-Instructions \citep{mishra2021crosstask}. SuperGLUE focuses on discriminative tasks, and Natural-Instructions is a set of generative tasks.

Empirically, we experiment with pretrained language models of two sizes: 125 million parameters and 1.3 billion parameters. We show that in our best setting, the 1.3 billion parameters model trained by the self-supervision performs better than the initial pretrained language models and two strong baselines on average.

Further analysis reveals that (1) the effectiveness of the self-supervision depends on the amount of training data, but the benefit of adding more data is diminishing; (2) the improvements brought by the self-supervision are in part due to the semantic similarity between the training and evaluation tasks; (3) adding more self-supervised objectives may not help model performance because adding them does not contribute to the diversity of the self-supervised tasks; (4) choosing similar task templates for both self-supervised and downstream tasks plays a vital role in improving model performance; (5) self-supervised tasks and human-annotated datasets are complementary; (6) generation examples show that compared to the initial pretrained language models, self-supervised-trained models are better at following the task instructions.

%% file: related.tex
\section{Related Work}
\paragraph{In-Context Few-Shot Learning.} \citet{gpt3} discover that large pretrained language models can solve unseen tasks at inference time. Recent work has improved the in-context few-shot performance by rescoring generated outputs \citep{pmlr-v139-zhao21c}, selecting \citep{liu2021makes} and ordering \citep{lu2021fantastically} the given few-shot examples.
Other work studies pretrained language models' cross-task generalization abilities for in-context few-shot or zero-shot learning using human-annotated datasets \citep{ye2021crossfit,wei2021finetuned,sanh2021multitask,min2021metaicl,xu2022zeroprompt} via instructions \citep{weller-etal-2020-learning,efrat2020turking,mishra2021crosstask,ouyang2022training} and retrieved examples \citep{hu2022context,lin2022unsupervised}. Our work differs in that we focus on self-supervised training.

\paragraph{Fine-Tuning for Few-Shot Learning.}
Pretrained language models for few-shot learning typically follows the ``pretrain then fine-tune'' paradigm \citep[\emph{inter alia}]{howard-ruder-2018-universal,radford2018improving,devlin-etal-2019-bert}, where recent work has focused on designing templates for few-shot fine-tuning \citep{reynolds2021prompt,schick-schutze-2021-exploiting,schick-schutze-2021-just,schick-schutze-2021-shot,le-scao-rush-2021-many,tam2021improving,gao-etal-2021-making,sorensen2022informationtheoretic}, and optimizing soft prompts \citep{li-liang-2021-prefix,qin-eisner-2021-learning,lester2021power,gu2021ppt,zhang2022differentiable}.
Other work focuses on unifying task formats to maximize the benefits of human annotations, including question answering \citep{leeadapting}, textual entailment \citep{yin-etal-2019-benchmarking,yin-etal-2020-universal,wang2021entailment}, and many other tasks \citep{McCann2018decaNLP,keskar2019unifying,2020t5,NEURIPS2021_8493eeac}.
In contrast, our focus is on in-context few-shot learning, without fine-tuning models on downstream task examples.

\paragraph{Pretraining for Few-Shot Learning.}
Several papers have adapted various resources for pretraining models to enhance their performances on few-shot learning, such as pretraining on hypertext \citep{aghajanyan2021htlm}, question-infused pretraining \citep{jia2021question}, and self-training \citep{du-etal-2021-self,vu-etal-2021-strata,wang2021list}. Pretraining approaches have targeted specific tasks, such as task-oriented dialog \citep{mi2021self}, intent detection \citep{zhang2021few}, and data-to-text generation \citep{chen-etal-2020-kgpt}. Our work differs as we use plain text as opposed to (naturally-occurring) human-annotated resources. Relatedly, \citet{bansal-etal-2020-self} used self-supervised meta-learning for few-shot text classification rather than in-context few-shot learning.

\paragraph{Intermediate Fine-Tuning.} Since our approach involves an extra training stage between pretraining and downstream evaluation, it is also related to prior work that uses multi-stage fine-tuning on human-annotated datasets for generic tasks \citep{phang2018sentence,pruksachatkun-etal-2020-intermediate,chang-lu-2021-rethinking-intermediate,aghajanyan-etal-2021-muppet,poth-etal-2021-pre} and text classification \citep{zhang-zhang-2021-qa}. Relevant work also studies intermediate fine-tuning using crosslingual supervision \citep{phang-etal-2020-english,moghe-etal-2021-cross}. \citet{rubino-sumita-2020-intermediate} use an intermediate self-supervised training stage for machine translation quality estimation.

%% file: method.tex
\section{Method}
\label{sec:method}
We describe four self-supervised training objectives that will be used to train models before downstream evaluations.

\begin{figure*}[t]
    \centering
    \includegraphics[scale=0.4]{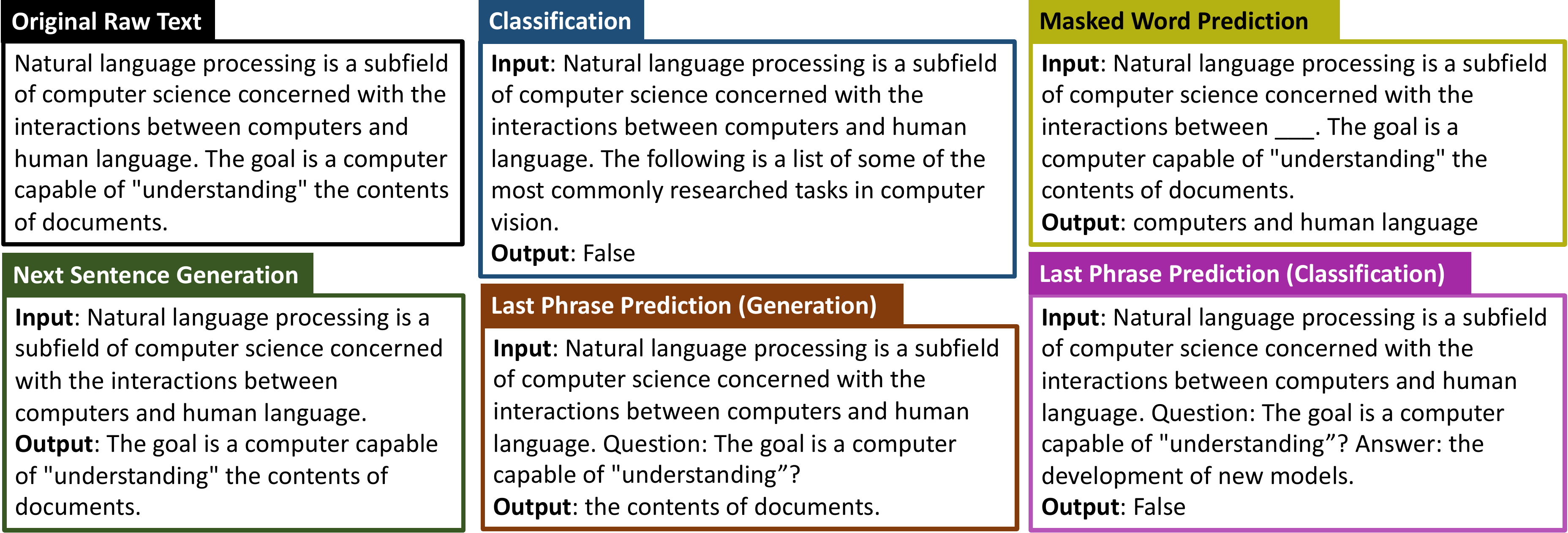}
    \caption{Examples of our self-supervised training tasks. Each example is an input-output pair constructed from the raw text.}
    \label{fig:selfsupervised_task_examples}
\end{figure*}

\begin{figure*}[t]
    \centering
    \includegraphics[scale=0.33]{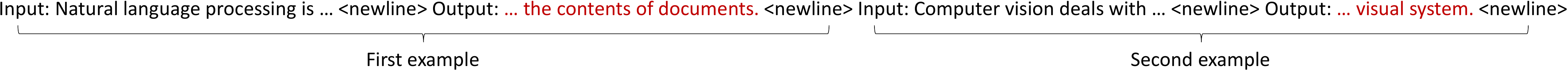}
    \caption{An example of a training instance. Each instance is formed by several training examples. During training, we use left-to-right language models and compute a cross-entropy loss on the output texts (indicated by the red color in the shown example). We note that when computing the loss on the second example, the first example can be seen as task demonstrations. For brevity, we show part of the input and output texts.}
    \label{fig:selfsupervised_instance_examples}
\end{figure*}

We begin by defining the example and the instance used during our self-supervised training. An \textbf{example} is an input-output pair. To differentiate the input and the output, we append special tokens ``Input:'' and ``Output:'' to the beginning of input text and output text respectively where the two texts are also separated by the $\langle\text{newline}\rangle$ token (see Figure~\ref{fig:selfsupervised_task_examples} for examples).\footnote{We chose this special symbol because we always start the self-supervised training from a pretrained language model checkpoint.}

An \textbf{instance} is a linearized string formed by several examples from the same task (e.g., see Figure~\ref{fig:selfsupervised_instance_examples}). As we encode the text using causal attention, the examples closer to the beginning of input sequences can be seen as task demonstrations, resulting in efficient computation.

When constructing the training examples, we pick three or more consecutive sentences (depending on the minimum sequence length we enforce on the sentences) and then apply task-specific rules to automatically create training data. To form a training instance, we randomly select examples from the same task until reaching the maximum sequence length (i.e., 2048). During training, we compute a cross-entropy loss on tokens in the \textbf{output texts}. We describe details of the self-supervised tasks in the following subsections.

\begin{figure*}
    \centering
    \includegraphics[scale=0.33]{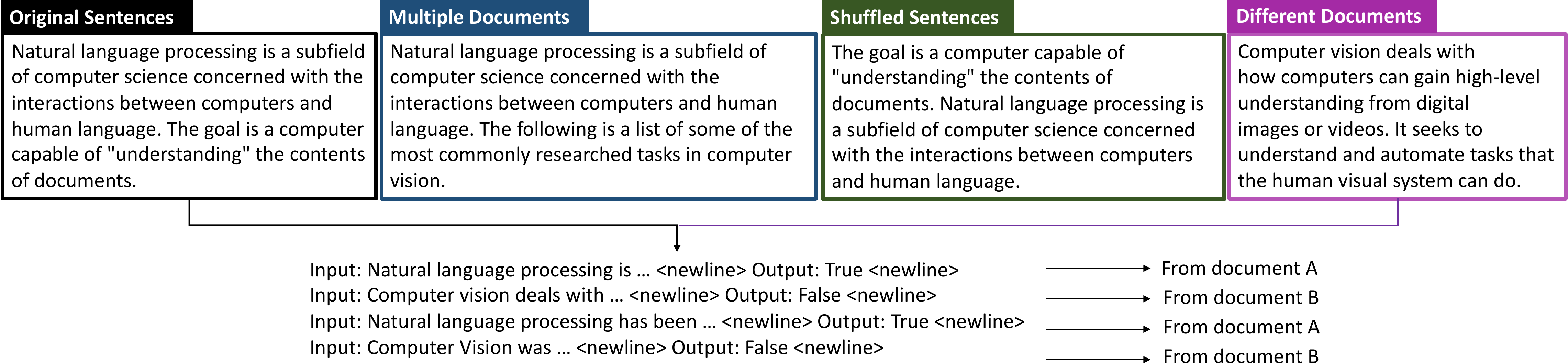}
    \caption{Example illustrating the construction of training instances for our classification task. There are four input types, and each training instance has two or three types. As the shown instance has the following two types: "original sentences" and "different documents", it comprises examples from two different documents. The instance resembles the next sentence prediction task, encouraging models to compare topical similarities between the two examples.}
    \label{fig:selfsupervised_classification_example}
    \vspace{-1em}
\end{figure*}

\subsection{Next Sentence Generation}

In light of the strong performance of language models on in-context few-shot learning \citep{gpt3}, we incorporate the language modeling as one of our self-supervised tasks, which we call ``next sentence generation'' (NSG). NSG asks the model to generate the next sentence given previous sentences as context. When building data for this task, we use the last sentence as output and the rest of the sentences as input.

\subsection{Masked Word Prediction}
The second task we consider is based on masked word prediction (MWP) which is commonly used in pretraining generic text encoders \citep{devlin-etal-2019-bert,liu2019roberta}. The task asks the model to fill in the missing information based on the surrounding context. Specifically, MWP randomly replaces words in input sentences with a special symbol and requires models to recover the masked words in the input. For this task, we create input text by randomly replacing 1{\raise.17ex\hbox{$\scriptstyle\mathtt{\sim}$}}20 words in the input text with a special token\footnote{We randomly select the special token from the following list: \_\_\_, $\langle\langle\rangle\rangle$, @@@, (()), \$\$\$, \%\%\%, \#\#\#, ***, and +++. We use random symbols instead of a fixed symbol because we found that it gives better performance in our preliminary experiments.} and use the masked out words as the output text.

\subsection{Last Phrase Prediction}
Inspired by the \lambada dataset \citep{paperno-etal-2016-lambada}, a question answering dataset which asks models to predict the last word in a sentence given several sentences of context, we create a ``last phrase prediction'' (\lpp) task, which requires predicting the last phrase in a sentence. To solve this task, models need to draw relevant information from the context and the learned knowledge during pretraining. We cast \lpp as either a generation task or a classification task. The latter variant of \lpp is a binary classification task that labels if the given answer is the correct phrase. To facilitate a unified format of these two tasks, we append a special token ``Question:'' to the beginning of the last sentence and replace the last phrase with a question mark. For the classification \lpp, we separate the given answer and the previous context and sentences with a special token ``Answer:''. An example of this task is shown in Figure \ref{fig:selfsupervised_task_examples}.

More specifically, we identify the last phrase of a sentence based on a set of function words (see appendix~\ref{appendix:lpp_function_words} for the list of function words). If there are multiple function words in a sentence, we pick the last one. Then we treat the text segment starting from the function word as the last phrase.\footnote{We ensure that the last sentence in raw text for this task always has at least one valid function word and the function word lies at the second half of the sentence.} When selecting negative answers, we randomly choose from the phrases extracted from the same function words (to make the negative answers more challenging).

\subsection{Classification}

Similar to the next sentence prediction task \citep{devlin-etal-2019-bert} and the sentence ordering prediction task \citep{jernite2017discourse,chen-etal-2019-evaluation} for pretraining language representations, we create a classification task (CL) for our self-supervised training. As shown in Figure \ref{fig:selfsupervised_classification_example}, for this task, we consider four types of input: original sentences, shuffled sentences, sentences from a different document, and sentences from multiple documents. In particular, for original sentences, we directly use text from original human-written documents. For shuffled sentences, we randomly shuffle all the input sentences. For sentences from multiple documents, we randomly replace 50\% of the input sentences with sentences from another document. We also ensure that the selected sentences (from both the input and another document) are consecutive in their original documents. For sentences from different documents, we replace the input sentences with sentences from another document. See Figure \ref{fig:selfsupervised_classification_example} for an example of each type of input.

When constructing a training instance, we randomly pick one or two additional input types and combine them with the original sentences to form a binary or three-way classification task. We also randomly assign label strings to input types in each instance to ensure that models follow the information given by earlier examples when making predictions.

The classification task is different from the other self-supervised tasks described in earlier subsections. It explicitly requires models to compare inputs across examples in a training instance to determine if the given input shares similar properties with the others. 

%% file: experiment.tex
\section{Experiment}

\begin{table*}[t]
    \centering\small
    \begin{tabular}{|l|l|}\hline
        \bf GPT3 & \$\{Context\}$\langle$newline$\rangle$\$\{Question\}$\langle$newline$\rangle$ - \bf\textcolor{red}{[\$\{Label\}] \$\{Answer\}} \\\hline
        \bf Ours & Input: \$\{Context\} Question: \$\{Question\} Answer: \$\{Answer\}$\langle$newline$\rangle$Output: \bf\textcolor{red}{\$\{Label\}} \\ \hline
    \end{tabular}
    \caption{Evaluation templates for MultiRC. \$\{$\cdot$\} represents values drawn from a particular data field. We alter the GPT3 template for this task to share a similar format with one of our self-supervised tasks (i.e., classification LLP in this case). The red, boldfaced texts are used to compute the language modeling perplexities for ranking the labels. We note that the shown template is for a single example, and there could be multiple examples within an instance.}
    \vspace{-1em}
    \label{tab:multirc_template}
\end{table*}

\subsection{Training Setup}

For the pretrained language model checkpoints, we use the 125 million parameters (125M) and the 1.3 billion parameters (1.3B) dense model from \citet{artetxe2021efficient}. These pretrained models have shown results comparable to GPT3 across various tasks.

For self-supervised training, we use a subset of documents from the RoBERTa training corpus \citep{liu2019roberta} that contains four domains: \textsc{BookCorpus} plus Wikipedia, \textsc{CC-News}, \textsc{OpenWebText}, and \textsc{Stories}. Specifically, we randomly sample 100k documents from each domain except \textsc{Stories} where we only sample 10k documents as the documents there are much longer than the others. The final training data contains approximately 1 million instances with 250k training instances per task.\footnote{The average numbers of example per instance for each data source are: 6.9 for \textsc{BookCorpus} plus Wikipedia, 5.3 for \textsc{CC-News}, 3.5 for \textsc{OpenWebText}, and 7.2 for \textsc{Stories}.} For the 125M model, we train for 10 epochs, which takes roughly 1 day on a V100 GPU. For the 1.3B model, we train for 5 epochs, which takes roughly 3 days on 2 V100 GPUs.

\subsection{Evaluation Setup}

The instance and example during evaluation shares similar definition as those in Sec.~\ref{sec:method} except that each evaluation instance has only one example from test splits and it is placed at the last position in the instance. The other examples in the instance (i.e., task demonstrations) come from either training splits or task-specific instructions depending on benchmarks.

We evaluate the models on two benchmarks: SuperGLUE and Natural-Instructions. SuperGLUE is a set of tasks focusing on natural language understanding. We use BoolQ (BQ; \citealp{clark-etal-2019-boolq}), CB \citep{demarneffe:cb}, COPA (CA; \citealp{roemmele2011choice}), MultiRC (MC; \citealp{khashabi-etal-2018-looking}), and RTE (RE; \citealp{giampiccolo-etal-2007-third,bentivogli2009fifth,dagan2006pascal,bar2006second}).\footnote{We exclude WSC \citep{levesque2011winograd} and ReCoRD \citep{zhang2018record} as pretrained models, including GPT3, require scoring algorithms at inference time to achieve competitive results. We exclude WiC \citep{pilehvar-camacho-collados-2019-wic} because GPT3-like models, including GPT3 and our models, do not give accuracies significantly better than random baselines.} We report results for the official development sets. The task demonstrations are examples randomly selected from the training sets. We report mean and standard deviations of five runs with different random seeds. Following GPT3, we use a ranking based approach when evaluating the models (i.e., pick the best label based on language modeling perplexities).

\paragraph{Natural-Instructions.} Natural-Instructions evaluates models' cross-task generalization abilities where all the tasks are generation tasks. It splits the tasks into two groups for training and evaluation. We use the same task split and evaluate models on the following task categories: question generation (QG), answer generation (AG), minimal modification (MM), and verification (VF).\footnote{We discard training tasks that share the same source datasets with evaluation tasks as we found that tasks with the same source dataset may contain leaked labels. We exclude the binary classification tasks because the class labels are severely imbalanced (i.e., more than 80\% of the class labels belong to one category).} Each task category has two tasks. Following the few-shot setting used in \citet{mishra2021crosstask}, we evaluate models using 100 examples per task, use greedy decoding, and report ROUGE-L \citep{lin-2004-rouge} scores per task category. For task demonstrations, we use the positive examples in the instructions in Natural-Instructions.

\begin{table*}[t]
    \centering\small
    \begin{tabular}{|l|c|c|c|c|c|c|c|}\hline
       \bf Model &\bf MS  &\bf BoolQ & \bf MultiRC &\bf COPA &\bf RTE &\bf CB &\bf Avg. \\\hline
       LM & 125M & 52.1(1.7) & 5.2(0.7)/49.5(1.1) & \bf 67.6(2.3) & 52.0(1.2) & 50.7(3.2)/34.8(2.5) & 48.4 \\
        ExtraLM & 125M & 51.5(1.7) & 5.1(0.8)/49.7(1.0) & 68.0(1.6) & 52.3(1.2) & 49.5(4.6)/35.5(5.6) & 48.3 \\
       NewTemplate & 125M & 52.2(1.8) & 5.2(0.6)/47.9(1.4) & 63.0(2.5) & 50.8(2.0) & 46.4(7.3)/30.1(6.4) & 46.2\\\hline
       CrossTask(NLI$\rightarrow$QA) & 125M & 38.1(0.3) & 5.1(0.7)/43.5(2.5) & 65.4(2.1) & - & - & \multirow{2}{*}{42.2} \\
        CrossTask (QA$\rightarrow$NLI) & 125M & - & - & - & \bf 53.6(0.5) & 39.6(1.5)/19.9(1.2) & \\\hline
        SameTask & 125M & 71.2 & 19.9/66.9 & 72.0 & 67.3 & 71.4/60.2 & 61.9 \\
        Self-Supervised & 125M & \bf 55.7(0.6) & \bf 7.0(1.0)/60.2(0.3) & \bf 67.6(2.1) & 53.0(1.5) & \bf 50.0(5.2)/39.8(3.0) & \bf 51.0 \\\hline\hline
        LM & 1.3B & 48.6(2.3) & 5.5(0.5)/53.7(0.7) & 83.4(1.7) & 51.9(1.2) & 53.6(5.2)/37.2(3.7) & 51.8 \\
        ExtraLM & 1.3B & 49.6(1.9) & 4.9(0.6)/54.8(0.6) & 82.6(1.5) & 52.9(1.9) & 51.4(7.5)/35.6(5.3) & 51.7  \\
        NewTemplate & 1.3B & 51.3(1.3) & 5.0(0.4)/52.8(1.2) & 81.2(2.4) & 50.8(2.3) & 49.3(4.7)/33.7(4.2) & 50.7 \\\hline
       CrossTask(NLI$\rightarrow$QA) & 1.3B & 53.4(0.8) & 1.2(0.3)/57.2(0.3) & 76.2(2.9) & - & - & \multirow{2}{*}{49.6} \\
        CrossTask (QA$\rightarrow$NLI) & 1.3B & - & - & - & \bf 54.3(1.2) & 44.6(3.6)/25.2(4.9) & \\\hline
        SameTask & 1.3B & 77.1 & 27.5/71.6 & 85.0 & 68.1 & 75.2/64.3 & 69.9 \\
        Self-Supervised & 1.3B & \bf 61.7(0.3) & \bf 5.2(0.1)/62.1(0.3) & \bf 84.0(2.7) & 53.1(0.7) & \bf 54.3(2.0)/37.0(1.9) & \bf 55.6 \\\hline
    \end{tabular}
    \caption{SuperGLUE results. We report mean and standard deviations (the numbers in parenthesis) of five runs. The best result (we take the average if there are two metrics) except SameTask in each column for each model size is boldfaced. MS=model size.}
    \label{tab:main_result_superglue}
\end{table*}

\begin{table}[t]
    \centering\small
    \begin{tabular}{|l|c|c|c|c|c|c|}\hline
        \bf Model & \bf MS & \bf QG & \bf AG & \bf MM & \bf VF & \bf Avg. \\\hline
        GPT3 & - & 43.0 & 50.0 & 70.0 & 32.0 & 48.8 \\\hline\hline
        LM & 125M & 33.7 & 12.9 & 53.0 & 14.7 & 28.6 \\ 
        ExtraLM & 125M &\bf 34.4 & 13.4 & 53.7 & 14.3 & 28.9 \\
        CrossTask & 125M & 22.0 & \bf 24.8 & 66.9 & 17.9 & \bf 32.9 \\
        SameTask & 125M & 54.8 & 42.3 & 77.3 & 78.3 & 63.2 \\
        SelfSup. & 125M & 16.9 & 14.6 & \bf 70.1 & \bf 18.9 & 30.0 \\\hline\hline
        LM & 1.3B & 40.9 & 32.5 & 74.0 & 27.8 & 43.8 \\ 
        ExtraLM & 1.3B & 41.1 & 32.7 & \bf 75.9 & 25.2 & 43.7 \\
        CrossTask & 1.3B & 38.1 & 41.6 & 69.2 & 23.0 & 42.9\\
        SameTask & 1.3B & 55.5 & 64.6 & 81.0 & 80.4 & 70.4\\
        SelfSup. & 1.3B & \bf 43.9 & \bf 37.5 & 72.3 & \bf 28.6 & \bf 45.5 \\\hline
    \end{tabular}
    \caption{Natural-Instructions results. The results for GPT3 are taken from \citet{mishra2021crosstask}. The best result except SameTask in each column for each model size is boldfaced. MS=model size.}
    \label{tab:main_result_naturalinstructions}
\end{table}

\paragraph{SuperGLUE.}
As our self-supervised tasks are formatted as input-output pairs, we change the task-specific templates for SuperGLUE to make them more similar to our self-supervised tasks. For example, as shown in Table~\ref{tab:multirc_template}, we make MultiRC similar to the classification LPP. More details of the template changes are in appendix~\ref{appendix_sec:superglue_templates}.

For both benchmarks, we also report an averaged performance for each model. For SuperGLUE, the average performance is computed based on the means of task performances. When a task has two metrics, we take the average of the two as the task performance.

More details on the dataset statistics and metrics for each task for both benchmarks are in appendix~\ref{sec:dataset_statistics}.

\paragraph{Baselines.} We consider four baselines: (1) directly evaluating pretrained language models on the benchmarks (LM) ; (2) performing additional language modeling training on the subset of the original data that is used for constructing the self-supervised tasks (ExtraLM). We use ExtraLM to approximately measure the contribution of additional computation; (3) fine-tuning on training sets for the tasks outside the evaluation sets (CrossTask). We use CrossTask to estimate the performances of cross-task supervision from human-annotated datasets; and (4) fine-tuning on training sets for the tasks in the evaluation sets (SameTask). SameTask serves as an oracle baseline estimating the approximated upperbound performances of cross-task supervision.

Since SuperGLUE does not have an official split for the CrossTask setting, we split the datasets into two groups according to the task category and report the CrossTask results based on ``CrossTask (QA$\rightarrow$NLI)'' and ``CrossTask (NLI$\rightarrow$QA)''.\footnote{``QA$\rightarrow$NLI'' suggests that we train models on the NLI tasks and evaluate on the QA tasks. Similarly, for ``NLI$\rightarrow$QA'', we train models on the QA tasks and evaluate on the NLI tasks.} As we alter the task templates, we report results for evaluating the pretrained language model checkpoints using the new templates (NewTemplate) to study the effect of new templates.

\subsection{Results}

\begin{table}[t]
    \centering\small
    \begin{tabular}{|l|c|c|c|c|c|c|}\hline
        \bf Model &\bf BQ &\bf MC &\bf CA &\bf RE &\bf CB &\bf Avg. \\\hline
         LM & 52.2 & 26.6 & 63.0 & 50.8 & 38.3 & 46.2 \\
         SelfSup. & 55.7 & 33.6 & 67.6 & 53.0 & 44.9 & 51.0\\\hline
         NSG & 52.1 & 25.9 & 64.0 & 51.0 & 41.2 & 46.9 \\
         CL & 52.5 & 26.8 & 61.4 & 50.9 & 48.1 & 47.9 \\
         MWP & 51.9 & 26.3 & 61.8 & 50.8 & 36.1 & 45.4 \\
         LPP & 53.5 & 29.5 & 61.6 & 52.0 & 40.3 & 47.4 \\\hline
    \end{tabular}
    \caption{SuperGLUE results when training the 125M model with only one of the self-supervised tasks.}
    \label{tab:effect_selfsupervised_superglue}
\end{table}

We report the results for SuperGLUE and Natural-Instructions in Table \ref{tab:main_result_superglue} and Table \ref{tab:main_result_naturalinstructions}. Our findings are as follows:
\begin{enumeratesquish}
\item Our proposed self-supervised training achieves the best performance on average for both benchmarks.
\item ExtraLM and NewTemplate show similar performances as the pretrained language model checkpoints, suggesting that the improvements from our self-supervised training is unlikely to come from the additional training on the data and the task template changes.
\item Compared to the pretrained language model checkpoints, CrossTask shows worse performances on both benchmarks, which is likely due to
the differences between training tasks and evaluation tasks.
\end{enumeratesquish}

%% file: analysis.tex
\section{Analysis}

\subsection{Effect of Amount of Data}

\begin{figure}[t]
    \centering\small
    \includegraphics[scale=0.3]{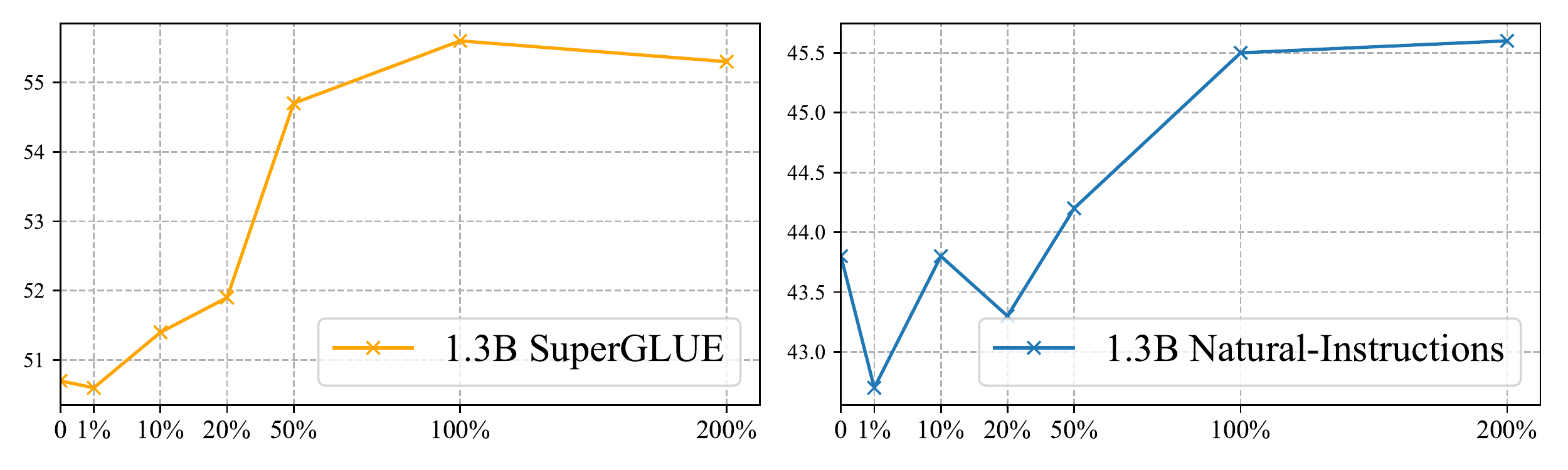}
    \caption{Average results for the 1.3B model on SuperGLUE and Natural-Instructions when varying the number of examples used for self-supervised training.}
    \label{fig:data_ratio}
\end{figure}

\begin{table}[t]
    \centering\small
    \begin{tabular}{|l|c|c|c|c|c|}\hline
        \bf Model & \bf QG & \bf AG & \bf MM & \bf VF & \bf Avg.  \\\hline
        LM & 33.7 & 12.9 & 53.0 & 14.7 & 28.6 \\
        SelfSup. & 16.9 & 14.6 & 70.1 & 18.9 & 30.0 \\\hline
        NSG & 32.3 & 12.5 & 54.0 & 13.8 & 28.2 \\
        CL & 8.3 & 0.3 & 1.0 & 2.7 & 3.1 \\
        MWP & 15.2 & 19.4 & 50.5 & 17.8 & 25.7 \\
        LPP & 11.3 & 16.6 & 49.5 & 19.9 & 24.3 \\\hline
    \end{tabular}
    \caption{Natural-Instructions results when training the 125M model with only one of the self-supervised tasks.}\vspace{-1em}
    \label{tab:effect_selfsupervised_naturalinstructions}
\end{table}

In Figure \ref{fig:data_ratio}, we report model performances for the 1.3B model on SuperGLUE and Natural-Instructions with 1\%, 10\%, 20\%, 50\%, and 200\% of training examples.\footnote{We apply the same ratio to all the self-supervised tasks and use the same development sets for each task across these settings.} We train the models for ten epochs.\footnote{Upon manual inspection, we found that the development set loss values in these experiments have converged.} As shown in the figure, when the amount of training data for self-supervised tasks is similar to that for the CrossTask setting (i.e., 1\% data), the self-supervised tasks also lead to worse performances. The improvements become clearer when we increase the number of training data, but it begins to plateau at around 100\% data. This suggests that one of the advantages of the self-supervised tasks compared to the tasks in the CrossTask setting is the amount of training data. We hypothesize that further increasing the amount of data not being helpful is because the data used for constructing the self-supervised tasks has already been used for language model pretraining. So, our models manage to learn to solve these tasks with a relatively limited amount of data. We have similar observations for the 125M model. See appendix~\ref{sec:effect_amount_data_full} for more details.\footnote{Our goal for this analysis is to show the rough trends of model performance when varying the amount of training data, rather than to provide an exact estimate of the training data required for the self-supervised training.}

\subsection{Effect of Individual Self-Supervised Tasks}

\begin{table}[t]
    \centering\small
    \begin{tabular}{|l|c|c|}\hline
         & \bf ALL & \bf ALL+MoreTask \\\hline
         \multicolumn{3}{|c|}{SuperGLUE Results}
         \\\hline
         125M & 51.0 & 50.9 \\
         1.3B & 55.6 & 55.6 \\\hline 
         \multicolumn{3}{|c|}{Natural-Instructions Results} \\\hline
         125M & 30.0 & 31.7 \\
         1.3B & 45.5 & 45.4 \\\hline 
    \end{tabular}
    \caption{Average results when adding denoising autoencoding and gap sentence prediction to the self-supervised training. ALL: use all of the self-supervision described in Sec.~\ref{sec:method}.}
    \label{tab:more_selfsupervised_tasks_results}
\end{table}
\begin{table}[t]
    \centering\small
    \begin{tabular}{|l|c|c|c|}\hline
         & \bf LM & \bf Correct Label & \bf Random Label \\\hline
         \multicolumn{4}{|c|}{SuperGLUE Results}
         \\\hline
         125M & 46.2 & 51.0 & 38.2 \\
         1.3B & 50.7 & 55.6 & 42.5 \\\hline 
         \multicolumn{4}{|c|}{Natural-Instructions Results} \\\hline
         125M & 28.6 & 30.0 & 19.1 \\
         1.3B & 43.8 & 45.5 & 31.5 \\\hline 
    \end{tabular}
    \caption{Average model performance comparing whether we assign random labels to the self-supervised tasks.}\vspace{-1em}
    \label{tab:corrupt_results}
\end{table}

\begin{table}[t]
    \centering\small
    \begin{tabular}{|l|c|c|c|}\hline
         & \bf MS & \bf GPT3 Template & \bf Our Template \\\hline
         LM & 125M & 48.4 & 46.2  \\
         SelfSup & 125M & 47.2 & 51.0 \\\hline
         LM & 1.3B & 51.8 & 50.7  \\
         SelfSup & 1.3B & 51.1 & 55.6 \\\hline 
    \end{tabular}
    \caption{Average results for SuperGLUE when using different task templates. MS=model size.}
    \label{tab:superglue_different_templates}
\end{table}

\begin{table}[t]
    \centering\small\setlength{\tabcolsep}{5pt}
    \begin{tabular}{|l|c|c|c|c|c|c|}\hline
        \bf Model & \bf MS & \bf QG & \bf AG & \bf MM & \bf VF & \bf Avg.  \\\hline
        LM & 125M & \bf 33.7 & 12.9 & 53.0 & 14.7 & 28.6 \\
        CrossTask & 125M & 22.0 & 24.8 & 66.9 & 17.9 & 32.9 \\
        SelfSup. & 125M & 16.9 & 14.6 & 70.1 & \bf 18.9 & 30.0 \\
        Combined & 125M & 23.5 & \bf 25.2 & \bf 70.3 & 18.5 & \bf 34.4 \\\hline\hline
        LM & 1.3B & 40.9 & 32.5 & 74.0 & 27.8 & 43.8 \\
        CrossTask & 1.3B & 38.1 & 41.6 & 69.2 & 23.0 & 42.9\\
        SelfSup. & 1.3B & \bf 43.9 & 37.5 & 72.3 & 28.6 & 45.5 \\
        Combined & 1.3B & 42.1 & \bf 42.5 & \bf 74.1 & \bf 28.7 & \bf 46.9 \\\hline
    \end{tabular}
    \caption{Natural-Instructions results when combining the self-supervised tasks and the tasks in the CrossTask setting. The best performance in each column for each model size is boldfaced. MS=model size.}\vspace{-1.3em}
    \label{tab:crosstask_selfsupervised_combine_naturalinstructions}
\end{table}

We investigate the effect of individual self-supervised tasks by training models with only one task. We report the experiment results in Table~\ref{tab:effect_selfsupervised_superglue} and Table~\ref{tab:effect_selfsupervised_naturalinstructions}. More results and discussions are in appendix~\ref{sec:effect_individual_task}. Our findings are:

\begin{enumeratesquish}
\item Combining all four self-supervised tasks results in the biggest improvements for most tasks, suggesting that the tasks are complementary.
\item Each self-supervised task improves a few downstream task performances (e.g., NSG helps COPA; CL helps MultiRC and CB). This is likely due to similarities between tasks.
\item It is worth noting that while CL hurts model performances on Natural-Instructions, it helps on the SuperGLUE. We hypothesis that this is because unlike Natural-Instructions, SuperGLUE is ranking based and, therefore, more favorable to classification-related training.
\item It is interesting to see that NSG and CL tasks are the two most beneficial to downstream performance among the four self-supervised tasks. This is likely due to (1) the generic task formulation of NSG, and (2) CL requires different inference abilities compared to the other self-supervised tasks. It is also interesting that training on only one of the self-supervised tasks can hurt the performance on Natural-Instruction.
\end{enumeratesquish}

\begin{table*}[t]
    \centering
    \small
    \begin{tabular}{|p{0.12\textwidth}|p{0.3\textwidth}|p{0.13\textwidth}|p{0.13\textwidth}|p{0.13\textwidth}|}\hline
        Task Prompt & Task Input & Reference & LM & Self-Supervised \\\hline
        Construct a question from the given fact by a simple rearrangement of words. & Fact: Pollen seeds come from male gametes of plants.
         & what seeds come from male gametes of plants? &  What might cause harm to plants? & What would you use to measure the number of male gametes of plants? \\\hline
        Ask a question on ``event duration'' based on the provided sentence. & Sentence: At the sight of the great man, Spear flushed crimson, and then his look of despair slowly disappeared; and into his eyes there came incredulously hope and gratitude. & How long did Spear see the great man? & How long did he stay in the Embassy? & How long did it take for Spear to look at the great man? \\\hline
        Answer the given question. Your answer must be a single span in the passage. & Passage: ... The following year he won a scholarship to the Royal Academy of Music, ... The principal of the Academy, Sir Alexander Mackenzie, had forbidden ... Question: What was the full name of the school Sir Alexander Mackenzie was principal of? & Royal Academy of Music. & Oliver. & the Royal Academy of Music.\\\hline
        Answer the given question. Your answer must be a single span in the passage. & Passage: ... Epitaph Records, founded by Brett Gurewitz of Bad Religion, was the base for many future pop punk bands ... The mainstream pop punk of latter-day bands such as Blink-182 is criticized by many punk rock devotees; in critic Christine Di Bella's words ... Question: What is the full name of the person that is very critical of modern mainstream pop punk bands? & Christine Di Bella. & the ``Bad Religion''. & many punk rock devotees. \\ \hline
    \end{tabular}
    \caption{Generation examples by the 1.3B model. The examples are taken from Natural-Instructions. The first two examples are from QG, and the other two are from AG. We only show part of the passages relevant to the outputs for QA for brevity.}\vspace{-1em}
    \label{tab:gen_examples}
\end{table*}

\subsection{Effect of More Self-Supervised Tasks}

To investigate the effect of having more self-supervised tasks during training, we add two extra self-supervised tasks to the self-supervised training, following the same procedure as the other tasks. The additional tasks are: denoising autoencoding \cite{lewis-etal-2020-bart} and gap sentence generation \cite{pmlr-v119-zhang20ae}. Denoising autoencoding is the task of reconstructing the original sentences from sentences corrupted by random noises, which has been shown effective for training generic language representations; gap sentence generation is to recover the missing sentence and has been found useful for abstractive summarization.

We report the results in Table \ref{tab:more_selfsupervised_tasks_results} where we do not find adding the two tasks improves downstream tasks. This is likely because the two tasks share similarities with our existing tasks (e.g., gap sentence generation shares a similar inference style as MWP). So, adding them does not promote diversity in the self-supervised tasks, leading to the fact that the models are not encouraged to learn different information.

\subsection{Effect of Few-Shot Templates}

The self-supervised training brings two benefits: making models familiar with the few-shot templates and task semantics. To differentiate the effect of the two, we train models on the self-supervised tasks with random labels. For example, for NSG, we use random sentences as outputs rather than the true next sentences; for the binary classification tasks, we randomly select binary labels. As shown in the results in Table \ref{tab:corrupt_results}, random labels hurt model performances, suggesting that what the models have learned is more than the few-shot templates.

We also investigate the effect of task templates for SuperGLUE by evaluating models using different templates. We report results in Table~\ref{tab:superglue_different_templates} where we find that having the templates for downstream tasks similar to the ones used for self-supervised training gives the models significantly better performances.

\subsection{Zero-Shot vs. One-Shot vs. Few-Shot}
\begin{table}[t]
    \centering\small
    \begin{tabular}{|l|c|c|c|c|c|c|}\hline
         & \multicolumn{2}{|c|}{\bf Zero-Shot} & \multicolumn{2}{|c|}{ \bf One-Shot} & \multicolumn{2}{|c|}{\bf Few-Shot} \\
             & LM & SS & LM & SS & LM & SS \\\hline
        125M & 46.7 & 44.3  & 42.6 & 46.1  & 46.2 & 51.0 \\\cline{2-7}
        \multicolumn{1}{|c|}{$\Delta$} &\multicolumn{2}{c|}{(-2.4)}  & \multicolumn{2}{c|}{(+3.5)} & \multicolumn{2}{c|}{\bf (+4.8)}  \\\hline
        1.3B & 49.5 & 49.9 & 46.5 & 50.8  & 50.7 & 55.6 \\\cline{2-7}
        \multicolumn{1}{|c|}{$\Delta$} & \multicolumn{2}{c|}{(+0.4)} & \multicolumn{2}{c|}{(+4.3)} & \multicolumn{2}{c|}{\bf (+4.9)} \\
        \hline
    \end{tabular}
    \caption{Average results for SuperGLUE showing the zero-shot, one-shot, and few-shot model performances for the LM and the self-supervised model (SS). The numbers in parenthesis are the performance differences between the LM and the SS with the positive numbers indicating improvements. We boldface the largest improvement for each model.}
    \label{tab:zeroshot_vs_oneshot_vs_fewshot}
\end{table}

We show zero-shot, one-shot, and few-shot performances for the LM and the self-supervised model in Table \ref{tab:zeroshot_vs_oneshot_vs_fewshot}. We find that among the three settings, the self-supervised training is the most helpful in the few-shot setting and does not help in the zero-shot setting, suggesting that the self-supervised training improves the models' in-context learning capabilities.

\subsection{Combine Self-Supervision with Cross-Task Human-Supervision}

We investigate the relations between the self-supervised tasks and the human-annotated tasks. We combine the tasks from the self-supervision and those from the CrossTask and report the results in Table \ref{tab:crosstask_selfsupervised_combine_naturalinstructions}. Interestingly, combining the two kinds of tasks results in better performances on average, showing that they are complementary.

\subsection{Generation Examples}

We show generation examples in Table \ref{tab:gen_examples}. In general, we find that compared to the vanilla pretrained language models, the self-supervised models are better at using information from task input following task requirements. Specifically, for the first two examples in Table \ref{tab:gen_examples}, the LM suffers from more severe semantic drift than the self-supervised model (e.g., ``male gametes of plants'' is more specific and relevant to the task input than ``plants''). We have similar observations for the third example, where ``Oliver'' is a name from the task demonstration rather than the passage. Interestingly, for the last example, the answer generated by the LM is from the passage but is actually ``the base for many future pop punk bands'' instead of what the question looks for (i.e., ``very critical of modern mainstream pop punk bands''). While the answer generated by the self-supervised model does not exactly match the reference, it is partially correct as the mainstream pop punk ``is criticized by many punk rock devotees''.

%% file: conclusion.tex
\section{Conclusion}
We evaluated four self-supervised objectives on two few-shot benchmarks by casting the self-supervised training as an intermediate training stage between language model pretraining and downstream few-shot evaluation. Empirically, we have shown that the models trained by the self-supervised objectives show the best performances compared to strong baselines on average. Analysis showed that (1) the amount of self-supervised training data and the diversity of the self-supervised tasks can affect the downstream performances.; (2) the self-supervised tasks are complementary to the human-annotated datasets; and (3) the self-supervised-trained models are better at following task requirements.

%% file: appendix.tex
\appendix
\section{Appendix}
\subsection{Additional Details for \lpp and Classification Tasks}

The label strings we used for LPP are as follows: Yes and No, Y and N, True and False, and T and F. We randomly choose from Yes, Y, True, and T as the label string for the positive label and use the other one in the selected pair as the negative label.

The label strings we used for the binary classification task are the same as the classification \lpp task. For the three-way classification task, we use the following label strings: Positive and Negative and Neutral, True and False and Neither, T and F and N, Yes and No and Unknown, Y and N and U.

\subsection{Dataset Statistics}
\label{sec:dataset_statistics}
\begin{table*}[t]
    \centering\small
    \begin{tabular}{|l|l|l|l|l|l|}\hline
        \bf Dataset & \bf Task Category & \bf Metrics & \bf\#Train & \bf\#Test & \bf\#Class \\\hline
        BoolQ &  Question Answering & Accuracy & 9427 & 3270 & 2 \\\hline
        MultiRC & Question Answering & F1$_a$/EM & 5100 & 953 & 2 \\\hline
        COPA & Question Answering & Accuracy & 400 & 100 & 2 \\\hline
        RTE & Natural Language Inference & Accuracy & 2500 & 278 & 2 \\\hline
        CB & Natural Language Inference & Accuracy/F1 & 250 & 57 & 3 \\\hline
    \end{tabular}
    \caption{Dataset statistics for SuperGLUE. We use the official development sets as test sets.}
    \label{tab:superglue_dataset_stat}
\end{table*}

\begin{table*}[t]
    \centering\small
    \begin{tabular}{|l|l|l|l|}\hline
        \bf Dataset & \bf Task Category & \bf \#Train & \bf\#Test \\\hline
        subtask003\_mctaco\_question\_generation\_event\_duration & Question Generation & 330 & 100  \\\hline
        subtask040\_qasc\_question\_generation & Question Generation & 6400 & 100 \\\hline
        subtask002\_quoref\_answer\_generation & Answer Generation & 6400 & 100 \\\hline
        subtask033\_winogrande\_answer\_generation & Answer Generation & 6400 & 100 \\\hline
        subtask034\_winogrande\_question\_modification\_object & Minimal Modification & 6400 & 100 \\\hline
        subtask045\_miscellaneous\_sentence\_paraphrasing & Minimal Modification & 93 & 100 \\\hline
        subtask039\_qasc\_find\_overlapping\_words & Verification & 6400 & 100 \\\hline
        subtask044\_essential\_terms\_identifying\_essential\_words & Verification & 2138 & 100 \\
        \hline
    \end{tabular}
    \caption{Dataset statistics for Natural-Instructions.}
    \label{tab:naturalinstructions_dataset_stat}
\end{table*}

We report dataset statistics for SuperGLUE and Natural-Instructions in Table~\ref{tab:superglue_dataset_stat} and Table~\ref{tab:naturalinstructions_dataset_stat}, respectively.

\subsection{Training Details}
\label{sec:train_details}

We train our models in PyTorch~\cite{paszke2017automatic} using \textsc{fairseq}~\citep{ott-etal-2019-fairseq}.

\subsection{More Details for Natural-Instructions}

\paragraph{Dataset Sources.}  CosmosQA \citep{huang-etal-2019-cosmos}, DROP \citep{dua-etal-2019-drop}, EssentialTerms \citep{khashabi-etal-2017-learning}, MCTACO \citep{zhou-etal-2019-going}, MultiRC \citep{khashabi-etal-2018-looking},
QASC \citep{Khot2020QASCAD}, Quoref \citep{dasigi-etal-2019-quoref}, ROPES \citep{lee-etal-2021-rope} and Winogrande \citep{Sakaguchi2020WINOGRANDEAA}.

\paragraph{Training Datasets.}
We used the following 8 datasets when training models in the cross-task setting: subtask026\_drop\_question\_generation, subtask060\_ropes\_question\_generation, subtask028\_drop\_answer\_generation, subtask047\_misc\_answering\_science\_questions, subtask061\_ropes\_answer\_generation, subtask059\_ropes\_story\_generation, subtask027\_drop\_answer\_type\_generation, subtask046\_miscellaenous\_question\_typing.

\subsection{List of Function Words for the Last Phrase Prediction Task}
\label{appendix:lpp_function_words}

We used the following function words for identifying the last phrase: the, a, an, for, including, and, in, is, are, were, was, neither, or, nor, be, at, in, on, by, to, would, will, before, after, of, about, from, excluding, except, during, under, above, then, into, onto, should, shall, must, may, might, than, with, using, can, could, about, as, from, within, without, have, had, been.

\subsection{Templates for SuperGLUE}
\label{appendix_sec:superglue_templates}
\begin{table*}[t]
    \centering\small
    \begin{subtable}{1\textwidth}
    \centering
    \begin{tabular}{|l|l|}\hline
        \bf GPT3 & \$\{Context\}$\langle$newline$\rangle$ question: \$\{Question\}$\langle$newline$\rangle$answer:\bf\textcolor{red}{\$\{Answer\}} \\\hline
        \bf Ours & Input: \$\{Context\} question: \$\{Question\} answer: True$\langle$newline$\rangle$Output: \bf\textcolor{red}{\$\{Answer\}} \\ \hline
    \end{tabular}
    \caption{BoolQ Template.}
    \end{subtable}\vspace{1em}
    \begin{subtable}{1\textwidth}
    \centering
    \begin{tabular}{|l|l|}\hline
        \bf GPT3 & \$\{Context\}$\langle$newline$\rangle$ question: \$\{Question\} True or False?$\langle$newline$\rangle$answer:\bf\textcolor{red}{\$\{Answer\}} \\\hline
        \bf Ours & Input: \$\{Context\} question: \$\{Question\} answer: True$\langle$newline$\rangle$Output: \bf\textcolor{red}{\$\{Answer\}} \\ \hline
    \end{tabular}
    \caption{RTE Template.}
    \end{subtable}\vspace{1em}
    \begin{subtable}{1\textwidth}
    \centering
    \begin{tabular}{|l|l|}\hline
        \bf GPT3 & \$\{Context\}$\langle$newline$\rangle$\bf\textcolor{red}{\$\{Answer\}} \\\hline
        \bf Ours & Input: \$\{Context\}$\langle$newline$\rangle$Output:\bf\textcolor{red}{\$\{Answer\}} \\ \hline
    \end{tabular}
    \caption{COPA Template.}
    \end{subtable}\vspace{1em}
    \begin{subtable}{1\textwidth}
    \centering
    \begin{tabular}{|l|l|}\hline
        \bf GPT3 & \$\{Context\}$\langle$newline$\rangle$ question: \$\{Question\} true, false, or neither?$\langle$newline$\rangle$answer:\bf\textcolor{red}{\$\{Answer\}} \\\hline
        \bf Ours & Input: \$\{Context\} question: \$\{Question\} true, false, or neither?$\langle$newline$\rangle$Output: \bf\textcolor{red}{\$\{Answer\}} \\ \hline
    \end{tabular}
    \caption{CB Template.}
    \end{subtable}
    \caption{Evaluation templates for SuperGLUE. \$\{$\cdot$\} represents values drawn from a particular data field. We alter the GPT3 templates for these tasks to share similar formats with one of our self-supervised tasks. The red, boldfaced texts are used to compute the language modeling perplexities for ranking the labels. We note that the shown templates are for a single example, and there could be multiple examples within an instance.}
    \label{appendix_tab:superglue_template}
\end{table*}

We show the SuperGLUE templates in Table~\ref{appendix_tab:superglue_template}.

\subsection{Hyperparameters}

We tune the hyperparameters based on development set performances. We tune the learning rate in \{1e-7, 5e-7, 1e-6, 3e-6, 5e-6, 8e-6, 1e-5, 3e-5, 5e-5\}, and the attention dropout rate in \{0.0, 0.1\}.

\subsection{Effect of Amount of Data}
\label{sec:effect_amount_data_full}

\begin{figure}[t]
    \centering\small
    \includegraphics[scale=0.3]{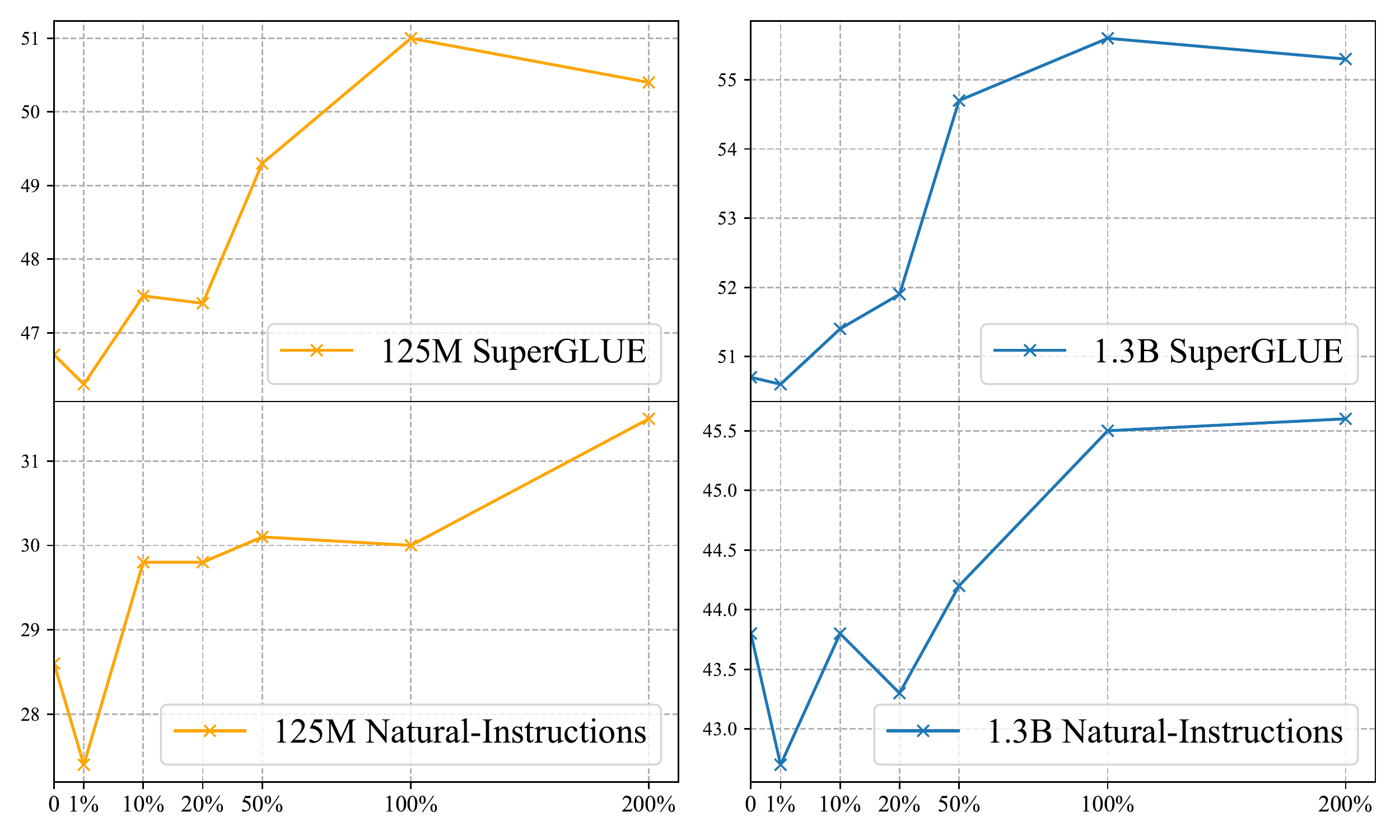}
    \caption{Average results on SuperGLUE and Natural-Instructions when varying number of examples used for training.}
    \label{fig:data_ratio_full}
\end{figure}

In Figure \ref{fig:data_ratio_full}, we report model performances for the 125M and 1.3B models on SuperGLUE and Natural-Instructions with 1\%, 10\%, 20\%, 50\%, and 200\% of training examples. 

\subsection{Effect of Individual Self-Supervised Task to Downstream Tasks}
\label{sec:effect_individual_task}

\begin{table}[t]
    \centering\small
    \begin{tabular}{|l|c|c|c|c|c|c|}\hline
        \bf Model &\bf BQ &\bf MC &\bf CA &\bf RE &\bf CB &\bf Avg. \\\hline
         LM & 52.2 & 26.6 & 63.0 & 50.8 & 38.3 & 46.2 \\
         SelfSup. & 55.7 & 33.6 & 67.6 & 53.0 & 44.9 & 51.0\\\hline
         NSG & 52.1 & 25.9 & 64.0 & 51.0 & 41.2 & 46.9 \\
         CL & 52.5 & 26.8 & 61.4 & 50.9 & 48.1 & 47.9 \\
         MWP & 51.9 & 26.3 & 61.8 & 50.8 & 36.1 & 45.4 \\
         LPP & 53.5 & 29.5 & 61.6 & 52.0 & 40.3 & 47.4 \\\hline
    \end{tabular}
    \caption{SuperGLUE results when training the 125M model with one of the self-supervised tasks.}
    \label{appendix_tab:effect_selfsupervised_superglue}
\end{table}

\begin{table}[t]
    \centering\small
    \begin{tabular}{|l|c|c|c|c|c|}\hline
        \bf Model & \bf QG & \bf AG & \bf MM & \bf VF & \bf Avg.  \\\hline
        LM & 33.7 & 12.9 & 53.0 & 14.7 & 28.6 \\
        SelfSup. & 16.9 & 14.6 & 70.1 & 18.9 & 30.0 \\\hline
        NSG & 32.3 & 12.5 & 54.0 & 13.8 & 28.2 \\
        CL & 8.3 & 0.3 & 1.0 & 2.7 & 3.1 \\
        MWP & 15.2 & 19.4 & 50.5 & 17.8 & 25.7 \\
        LPP & 11.3 & 16.6 & 49.5 & 19.9 & 24.3 \\\hline
    \end{tabular}
    \caption{Natural-Instructions results when training the 125M model with one of the self-supervised tasks.}
    \label{appendix_tab:effect_selfsupervised_naturalinstructions}
\end{table}

\begin{table}[t]
    \centering\small
    \begin{tabular}{|l|c|c|c|c|c|c|}\hline
        \bf Model &\bf BQ &\bf MC &\bf CA &\bf RE &\bf CB &\bf Avg. \\\hline
         LM & 52.2 & 26.6 & 63.0 & 50.8 & 38.3 & 46.2 \\
         ALL & 55.7 & 33.6 & 67.6 & 53.0 & 44.9 & 51.0\\\hline
         ALL-NSG & 55.0 & 31.8 & 62.7 & 52.5 & 45.9 & 49.6 \\
         ALL-MWP & 55.6 & 33.5 & 67.3 & 52.7 & 45.5 & 50.9 \\
         ALL-LPP & 53.5 & 30.5 & 67.6 & 51.9 & 46.6 & 50.0 \\
         ALL-CL & 54.0 & 32.9 & 67.4 & 52.8 & 39.0 & 49.2 \\
         \hline
    \end{tabular}
    \caption{SuperGLUE results when excluding one of the self-supervised tasks. The results are based on the 125M model.}
    \label{appendix_tab:effect_selfsupervised_superglue_exclude}
\end{table}
\begin{table}[t]
    \centering\small
    \begin{tabular}{|l|c|c|c|c|c|}\hline
        \bf Model & \bf QG & \bf AG & \bf MM & \bf VF & \bf Avg.  \\\hline
        LM & 33.7 & 12.9 & 53.0 & 14.7 & 28.6 \\
        ALL & 16.9 & 14.6 & 70.1 & 18.9 & 30.0 \\\hline
        ALL-NSG & 10.5 & 18.7 & 46.5 & 14.9 & 22.7\\
        ALL-MWP & 17.2 & 14.9 & 67.1 & 17.9 & 29.3\\
        ALL-LPP & 17.3 & 14.8 & 67.6 & 18.1 & 29.5 \\
        ALL-CL & 23.1 & 15.1 & 59.0 & 18.2 & 28.9 \\
        \hline
    \end{tabular}
    \caption{Natural-Instructions results when excluding one of the self-supervised tasks. The results are based on the 125M model.}
    \label{appendix_tab:effect_selfsupervised_naturalinstructions_exclude}
\end{table}

We investigate the effect of individual self-supervised task by considering two experiment settings: training models with only one task and training models with one task excluded. We report the experiment results in Table~\ref{appendix_tab:effect_selfsupervised_superglue}, Table~\ref{appendix_tab:effect_selfsupervised_naturalinstructions}, Table~\ref{appendix_tab:effect_selfsupervised_superglue_exclude}, and Table~\ref{appendix_tab:effect_selfsupervised_naturalinstructions_exclude}. Our findings are:

\begin{enumeratesquish}
\item Combining all the four self-supervised tasks gives the largest improvements for most tasks, suggesting that these tasks are mostly complementary.
\item Each self-supervised task improves a few downstream task performances (e.g., NSG helps COPA; CL helps MultiRC and CB). This is likely due to the semantic similarities between tasks.
\item It is worth noting that (1) while CL hurts model performances on Natural-Instructions, it helps on the SuperGLUE; and (2) excluding NSG hurts the model performances most on Natural-Instructions whereas excluding CL hurts the most on SuperGLUE. This presumably is because SuperGLUE is ranking based and therefore is more favorable to classification-related training, whereas the tasks in Natural-Instructions are generation tasks and thus benefits more from generation-related tasks.
\item It is interesting to see that among the four self-supervised tasks, NSG and CL tasks are the two most important factors in terms of affecting the downstream performances. This is likely due to (1) the generic task formulation of NSG and it being the only sentence generation tasks; and (2) the drastic differences between CL and the other self-supervised tasks with respect to their inference styles. Unlike NSG/MWP/LPP, which models can rely on input within each example to solve the task, CL require models to make comparisons across examples in a training instance.
\end{enumeratesquish}

%% file: acl_latex.bbl
\begin{thebibliography}{87}
\expandafter\ifx\csname natexlab\endcsname\relax\def\natexlab#1{#1}\fi

\bibitem[{Aghajanyan et~al.(2021{\natexlab{a}})Aghajanyan, Gupta, Shrivastava,
  Chen, Zettlemoyer, and Gupta}]{aghajanyan-etal-2021-muppet}
Armen Aghajanyan, Anchit Gupta, Akshat Shrivastava, Xilun Chen, Luke
  Zettlemoyer, and Sonal Gupta. 2021{\natexlab{a}}.
\newblock \href {https://doi.org/10.18653/v1/2021.emnlp-main.468} {Muppet:
  Massive multi-task representations with pre-finetuning}.
\newblock In \emph{Proceedings of the 2021 Conference on Empirical Methods in
  Natural Language Processing}, pages 5799--5811, Online and Punta Cana,
  Dominican Republic. Association for Computational Linguistics.

\bibitem[{Aghajanyan et~al.(2021{\natexlab{b}})Aghajanyan, Okhonko, Lewis,
  Joshi, Xu, Ghosh, and Zettlemoyer}]{aghajanyan2021htlm}
Armen Aghajanyan, Dmytro Okhonko, Mike Lewis, Mandar Joshi, Hu~Xu, Gargi Ghosh,
  and Luke Zettlemoyer. 2021{\natexlab{b}}.
\newblock Htlm: hyper-text pre-training and prompting of language models.
\newblock \emph{arXiv preprint arXiv:2107.06955}.

\bibitem[{Artetxe et~al.(2021)Artetxe, Bhosale, Goyal, Mihaylov, Ott, Shleifer,
  Lin, Du, Iyer, Pasunuru et~al.}]{artetxe2021efficient}
Mikel Artetxe, Shruti Bhosale, Naman Goyal, Todor Mihaylov, Myle Ott, Sam
  Shleifer, Xi~Victoria Lin, Jingfei Du, Srinivasan Iyer, Ramakanth Pasunuru,
  et~al. 2021.
\newblock Efficient large scale language modeling with mixtures of experts.
\newblock \emph{arXiv preprint arXiv:2112.10684}.

\bibitem[{Bansal et~al.(2020)Bansal, Jha, Munkhdalai, and
  McCallum}]{bansal-etal-2020-self}
Trapit Bansal, Rishikesh Jha, Tsendsuren Munkhdalai, and Andrew McCallum. 2020.
\newblock \href {https://doi.org/10.18653/v1/2020.emnlp-main.38}
  {Self-supervised meta-learning for few-shot natural language classification
  tasks}.
\newblock In \emph{Proceedings of the 2020 Conference on Empirical Methods in
  Natural Language Processing (EMNLP)}, pages 522--534, Online. Association for
  Computational Linguistics.

\bibitem[{Bar~Haim et~al.(2006)Bar~Haim, Dagan, Dolan, Ferro, Giampiccolo,
  Magnini, and Szpektor}]{bar2006second}
Roy Bar~Haim, Ido Dagan, Bill Dolan, Lisa Ferro, Danilo Giampiccolo, Bernardo
  Magnini, and Idan Szpektor. 2006.
\newblock The second {PASCAL} recognising textual entailment challenge.

\bibitem[{Bentivogli et~al.(2009)Bentivogli, Dagan, Dang, Giampiccolo, and
  Magnini}]{bentivogli2009fifth}
Luisa Bentivogli, Ido Dagan, Hoa~Trang Dang, Danilo Giampiccolo, and Bernardo
  Magnini. 2009.
\newblock The fifth {PASCAL} recognizing textual entailment challenge.

\bibitem[{Bragg et~al.(2021)Bragg, Cohan, Lo, and
  Beltagy}]{NEURIPS2021_8493eeac}
Jonathan Bragg, Arman Cohan, Kyle Lo, and Iz~Beltagy. 2021.
\newblock \href
  {https://proceedings.neurips.cc/paper/2021/file/8493eeaccb772c0878f99d60a0bd2bb3-Paper.pdf}
  {Flex: Unifying evaluation for few-shot nlp}.
\newblock In \emph{Advances in Neural Information Processing Systems},
  volume~34, pages 15787--15800. Curran Associates, Inc.

\bibitem[{Brown et~al.(2020)Brown, Mann, Ryder, Subbiah, Kaplan, Dhariwal,
  Neelakantan, Shyam, Sastry, Askell, Agarwal, Herbert-Voss, Krueger, Henighan,
  Child, Ramesh, Ziegler, Wu, Winter, Hesse, Chen, Sigler, Litwin, Gray, Chess,
  Clark, Berner, McCandlish, Radford, Sutskever, and Amodei}]{gpt3}
Tom Brown, Benjamin Mann, Nick Ryder, Melanie Subbiah, Jared~D Kaplan, Prafulla
  Dhariwal, Arvind Neelakantan, Pranav Shyam, Girish Sastry, Amanda Askell,
  Sandhini Agarwal, Ariel Herbert-Voss, Gretchen Krueger, Tom Henighan, Rewon
  Child, Aditya Ramesh, Daniel Ziegler, Jeffrey Wu, Clemens Winter, Chris
  Hesse, Mark Chen, Eric Sigler, Mateusz Litwin, Scott Gray, Benjamin Chess,
  Jack Clark, Christopher Berner, Sam McCandlish, Alec Radford, Ilya Sutskever,
  and Dario Amodei. 2020.
\newblock \href
  {https://proceedings.neurips.cc/paper/2020/file/1457c0d6bfcb4967418bfb8ac142f64a-Paper.pdf}
  {Language models are few-shot learners}.
\newblock In \emph{Advances in Neural Information Processing Systems},
  volume~33, pages 1877--1901. Curran Associates, Inc.

\bibitem[{Chang and Lu(2021)}]{chang-lu-2021-rethinking-intermediate}
Ting-Yun Chang and Chi-Jen Lu. 2021.
\newblock \href {https://doi.org/10.18653/v1/2021.findings-emnlp.61}
  {Rethinking why intermediate-task fine-tuning works}.
\newblock In \emph{Findings of the Association for Computational Linguistics:
  EMNLP 2021}, pages 706--713, Punta Cana, Dominican Republic. Association for
  Computational Linguistics.

\bibitem[{Chen et~al.(2019)Chen, Chu, and Gimpel}]{chen-etal-2019-evaluation}
Mingda Chen, Zewei Chu, and Kevin Gimpel. 2019.
\newblock \href {https://doi.org/10.18653/v1/D19-1060} {Evaluation benchmarks
  and learning criteria for discourse-aware sentence representations}.
\newblock In \emph{Proceedings of the 2019 Conference on Empirical Methods in
  Natural Language Processing and the 9th International Joint Conference on
  Natural Language Processing (EMNLP-IJCNLP)}, pages 649--662, Hong Kong,
  China. Association for Computational Linguistics.

\bibitem[{Chen et~al.(2020)Chen, Su, Yan, and Wang}]{chen-etal-2020-kgpt}
Wenhu Chen, Yu~Su, Xifeng Yan, and William~Yang Wang. 2020.
\newblock \href {https://doi.org/10.18653/v1/2020.emnlp-main.697} {{KGPT}:
  Knowledge-grounded pre-training for data-to-text generation}.
\newblock In \emph{Proceedings of the 2020 Conference on Empirical Methods in
  Natural Language Processing (EMNLP)}, pages 8635--8648, Online. Association
  for Computational Linguistics.

\bibitem[{Clark et~al.(2019)Clark, Lee, Chang, Kwiatkowski, Collins, and
  Toutanova}]{clark-etal-2019-boolq}
Christopher Clark, Kenton Lee, Ming-Wei Chang, Tom Kwiatkowski, Michael
  Collins, and Kristina Toutanova. 2019.
\newblock \href {https://doi.org/10.18653/v1/N19-1300} {{B}ool{Q}: Exploring
  the surprising difficulty of natural yes/no questions}.
\newblock In \emph{Proceedings of the 2019 Conference of the North {A}merican
  Chapter of the Association for Computational Linguistics: Human Language
  Technologies, Volume 1 (Long and Short Papers)}, pages 2924--2936,
  Minneapolis, Minnesota. Association for Computational Linguistics.

\bibitem[{Dagan et~al.(2006)Dagan, Glickman, and Magnini}]{dagan2006pascal}
Ido Dagan, Oren Glickman, and Bernardo Magnini. 2006.
\newblock The {PASCAL} recognising textual entailment challenge.
\newblock In \emph{Machine learning challenges. evaluating predictive
  uncertainty, visual object classification, and recognising tectual
  entailment}, pages 177--190. Springer.

\bibitem[{Dasigi et~al.(2019)Dasigi, Liu, Marasovi{\'c}, Smith, and
  Gardner}]{dasigi-etal-2019-quoref}
Pradeep Dasigi, Nelson~F. Liu, Ana Marasovi{\'c}, Noah~A. Smith, and Matt
  Gardner. 2019.
\newblock \href {https://doi.org/10.18653/v1/D19-1606} {{Q}uoref: A reading
  comprehension dataset with questions requiring coreferential reasoning}.
\newblock In \emph{Proceedings of the 2019 Conference on Empirical Methods in
  Natural Language Processing and the 9th International Joint Conference on
  Natural Language Processing (EMNLP-IJCNLP)}, pages 5925--5932, Hong Kong,
  China. Association for Computational Linguistics.

\bibitem[{De~Marneffe et~al.(2019)De~Marneffe, Simons, and
  Tonhauser}]{demarneffe:cb}
Marie-Catherine De~Marneffe, Mandy Simons, and Judith Tonhauser. 2019.
\newblock {The CommitmentBank}: Investigating projection in naturally occurring
  discourse.
\newblock To appear in proceedings of Sinn und Bedeutung 23. Data can be found
  at https://github.com/mcdm/CommitmentBank/.

\bibitem[{Devlin et~al.(2019)Devlin, Chang, Lee, and
  Toutanova}]{devlin-etal-2019-bert}
Jacob Devlin, Ming-Wei Chang, Kenton Lee, and Kristina Toutanova. 2019.
\newblock \href {https://doi.org/10.18653/v1/N19-1423} {{BERT}: Pre-training of
  deep bidirectional transformers for language understanding}.
\newblock In \emph{Proceedings of the 2019 Conference of the North {A}merican
  Chapter of the Association for Computational Linguistics: Human Language
  Technologies, Volume 1 (Long and Short Papers)}, pages 4171--4186,
  Minneapolis, Minnesota. Association for Computational Linguistics.

\bibitem[{Du et~al.(2021)Du, Grave, Gunel, Chaudhary, Celebi, Auli, Stoyanov,
  and Conneau}]{du-etal-2021-self}
Jingfei Du, Edouard Grave, Beliz Gunel, Vishrav Chaudhary, Onur Celebi, Michael
  Auli, Veselin Stoyanov, and Alexis Conneau. 2021.
\newblock \href {https://doi.org/10.18653/v1/2021.naacl-main.426}
  {Self-training improves pre-training for natural language understanding}.
\newblock In \emph{Proceedings of the 2021 Conference of the North American
  Chapter of the Association for Computational Linguistics: Human Language
  Technologies}, pages 5408--5418, Online. Association for Computational
  Linguistics.

\bibitem[{Dua et~al.(2019)Dua, Wang, Dasigi, Stanovsky, Singh, and
  Gardner}]{dua-etal-2019-drop}
Dheeru Dua, Yizhong Wang, Pradeep Dasigi, Gabriel Stanovsky, Sameer Singh, and
  Matt Gardner. 2019.
\newblock \href {https://doi.org/10.18653/v1/N19-1246} {{DROP}: A reading
  comprehension benchmark requiring discrete reasoning over paragraphs}.
\newblock In \emph{Proceedings of the 2019 Conference of the North {A}merican
  Chapter of the Association for Computational Linguistics: Human Language
  Technologies, Volume 1 (Long and Short Papers)}, pages 2368--2378,
  Minneapolis, Minnesota. Association for Computational Linguistics.

\bibitem[{Efrat and Levy(2020)}]{efrat2020turking}
Avia Efrat and Omer Levy. 2020.
\newblock The turking test: Can language models understand instructions?
\newblock \emph{arXiv preprint arXiv:2010.11982}.

\bibitem[{Gao et~al.(2021)Gao, Fisch, and Chen}]{gao-etal-2021-making}
Tianyu Gao, Adam Fisch, and Danqi Chen. 2021.
\newblock \href {https://doi.org/10.18653/v1/2021.acl-long.295} {Making
  pre-trained language models better few-shot learners}.
\newblock In \emph{Proceedings of the 59th Annual Meeting of the Association
  for Computational Linguistics and the 11th International Joint Conference on
  Natural Language Processing (Volume 1: Long Papers)}, pages 3816--3830,
  Online. Association for Computational Linguistics.

\bibitem[{Giampiccolo et~al.(2007)Giampiccolo, Magnini, Dagan, and
  Dolan}]{giampiccolo-etal-2007-third}
Danilo Giampiccolo, Bernardo Magnini, Ido Dagan, and Bill Dolan. 2007.
\newblock \href {https://aclanthology.org/W07-1401} {The third {PASCAL}
  recognizing textual entailment challenge}.
\newblock In \emph{Proceedings of the {ACL}-{PASCAL} Workshop on Textual
  Entailment and Paraphrasing}, pages 1--9, Prague. Association for
  Computational Linguistics.

\bibitem[{Gu et~al.(2021)Gu, Han, Liu, and Huang}]{gu2021ppt}
Yuxian Gu, Xu~Han, Zhiyuan Liu, and Minlie Huang. 2021.
\newblock Ppt: Pre-trained prompt tuning for few-shot learning.
\newblock \emph{arXiv preprint arXiv:2109.04332}.

\bibitem[{Howard and Ruder(2018)}]{howard-ruder-2018-universal}
Jeremy Howard and Sebastian Ruder. 2018.
\newblock \href {https://doi.org/10.18653/v1/P18-1031} {Universal language
  model fine-tuning for text classification}.
\newblock In \emph{Proceedings of the 56th Annual Meeting of the Association
  for Computational Linguistics (Volume 1: Long Papers)}, pages 328--339,
  Melbourne, Australia. Association for Computational Linguistics.

\bibitem[{Hu et~al.(2022)Hu, Lee, Xie, Yu, Smith, and
  Ostendorf}]{hu2022context}
Yushi Hu, Chia-Hsuan Lee, Tianbao Xie, Tao Yu, Noah~A Smith, and Mari
  Ostendorf. 2022.
\newblock In-context learning for few-shot dialogue state tracking.
\newblock \emph{arXiv preprint arXiv:2203.08568}.

\bibitem[{Huang et~al.(2019)Huang, Le~Bras, Bhagavatula, and
  Choi}]{huang-etal-2019-cosmos}
Lifu Huang, Ronan Le~Bras, Chandra Bhagavatula, and Yejin Choi. 2019.
\newblock \href {https://doi.org/10.18653/v1/D19-1243} {Cosmos {QA}: Machine
  reading comprehension with contextual commonsense reasoning}.
\newblock In \emph{Proceedings of the 2019 Conference on Empirical Methods in
  Natural Language Processing and the 9th International Joint Conference on
  Natural Language Processing (EMNLP-IJCNLP)}, pages 2391--2401, Hong Kong,
  China. Association for Computational Linguistics.

\bibitem[{Jernite et~al.(2017)Jernite, Bowman, and
  Sontag}]{jernite2017discourse}
Yacine Jernite, Samuel~R Bowman, and David Sontag. 2017.
\newblock Discourse-based objectives for fast unsupervised sentence
  representation learning.
\newblock \emph{arXiv preprint arXiv:1705.00557}.

\bibitem[{Jia et~al.(2021)Jia, Lewis, and Zettlemoyer}]{jia2021question}
Robin Jia, Mike Lewis, and Luke Zettlemoyer. 2021.
\newblock Question answering infused pre-training of general-purpose
  contextualized representations.
\newblock \emph{arXiv preprint arXiv:2106.08190}.

\bibitem[{Keskar et~al.(2019)Keskar, McCann, Xiong, and
  Socher}]{keskar2019unifying}
Nitish~Shirish Keskar, Bryan McCann, Caiming Xiong, and Richard Socher. 2019.
\newblock Unifying question answering, text classification, and regression via
  span extraction.
\newblock \emph{arXiv preprint arXiv:1904.09286}.

\bibitem[{Khashabi et~al.(2018)Khashabi, Chaturvedi, Roth, Upadhyay, and
  Roth}]{khashabi-etal-2018-looking}
Daniel Khashabi, Snigdha Chaturvedi, Michael Roth, Shyam Upadhyay, and Dan
  Roth. 2018.
\newblock \href {https://doi.org/10.18653/v1/N18-1023} {Looking beyond the
  surface: A challenge set for reading comprehension over multiple sentences}.
\newblock In \emph{Proceedings of the 2018 Conference of the North {A}merican
  Chapter of the Association for Computational Linguistics: Human Language
  Technologies, Volume 1 (Long Papers)}, pages 252--262, New Orleans,
  Louisiana. Association for Computational Linguistics.

\bibitem[{Khashabi et~al.(2017)Khashabi, Khot, Sabharwal, and
  Roth}]{khashabi-etal-2017-learning}
Daniel Khashabi, Tushar Khot, Ashish Sabharwal, and Dan Roth. 2017.
\newblock \href {https://doi.org/10.18653/v1/K17-1010} {Learning what is
  essential in questions}.
\newblock In \emph{Proceedings of the 21st Conference on Computational Natural
  Language Learning ({C}o{NLL} 2017)}, pages 80--89, Vancouver, Canada.
  Association for Computational Linguistics.

\bibitem[{Khot et~al.(2020)Khot, Clark, Guerquin, Jansen, and
  Sabharwal}]{Khot2020QASCAD}
Tushar Khot, Peter Clark, Michal Guerquin, Peter~Alexander Jansen, and Ashish
  Sabharwal. 2020.
\newblock Qasc: A dataset for question answering via sentence composition.
\newblock In \emph{AAAI}.

\bibitem[{Le~Scao and Rush(2021)}]{le-scao-rush-2021-many}
Teven Le~Scao and Alexander Rush. 2021.
\newblock \href {https://doi.org/10.18653/v1/2021.naacl-main.208} {How many
  data points is a prompt worth?}
\newblock In \emph{Proceedings of the 2021 Conference of the North American
  Chapter of the Association for Computational Linguistics: Human Language
  Technologies}, pages 2627--2636, Online. Association for Computational
  Linguistics.

\bibitem[{Lee et~al.(2021)Lee, Li, Wang, Wang, Fujii, Qin, Popat, and
  Pfister}]{lee-etal-2021-rope}
Chen-Yu Lee, Chun-Liang Li, Chu Wang, Renshen Wang, Yasuhisa Fujii, Siyang Qin,
  Ashok Popat, and Tomas Pfister. 2021.
\newblock \href {https://doi.org/10.18653/v1/2021.acl-short.41} {{ROPE}:
  Reading order equivariant positional encoding for graph-based document
  information extraction}.
\newblock In \emph{Proceedings of the 59th Annual Meeting of the Association
  for Computational Linguistics and the 11th International Joint Conference on
  Natural Language Processing (Volume 2: Short Papers)}, pages 314--321,
  Online. Association for Computational Linguistics.

\bibitem[{Lester et~al.(2021)Lester, Al-Rfou, and Constant}]{lester2021power}
Brian Lester, Rami Al-Rfou, and Noah Constant. 2021.
\newblock The power of scale for parameter-efficient prompt tuning.
\newblock \emph{arXiv preprint arXiv:2104.08691}.

\bibitem[{Levesque et~al.(2011)Levesque, Davis, and
  Morgenstern}]{levesque2011winograd}
Hector~J Levesque, Ernest Davis, and Leora Morgenstern. 2011.
\newblock The {W}inograd schema challenge.
\newblock In \emph{{AAAI} Spring Symposium: Logical Formalizations of
  Commonsense Reasoning}, volume~46, page~47.

\bibitem[{Lewis et~al.(2020)Lewis, Liu, Goyal, Ghazvininejad, Mohamed, Levy,
  Stoyanov, and Zettlemoyer}]{lewis-etal-2020-bart}
Mike Lewis, Yinhan Liu, Naman Goyal, Marjan Ghazvininejad, Abdelrahman Mohamed,
  Omer Levy, Veselin Stoyanov, and Luke Zettlemoyer. 2020.
\newblock \href {https://doi.org/10.18653/v1/2020.acl-main.703} {{BART}:
  Denoising sequence-to-sequence pre-training for natural language generation,
  translation, and comprehension}.
\newblock In \emph{Proceedings of the 58th Annual Meeting of the Association
  for Computational Linguistics}, pages 7871--7880, Online. Association for
  Computational Linguistics.

\bibitem[{Li and Liang(2021)}]{li-liang-2021-prefix}
Xiang~Lisa Li and Percy Liang. 2021.
\newblock \href {https://doi.org/10.18653/v1/2021.acl-long.353} {Prefix-tuning:
  Optimizing continuous prompts for generation}.
\newblock In \emph{Proceedings of the 59th Annual Meeting of the Association
  for Computational Linguistics and the 11th International Joint Conference on
  Natural Language Processing (Volume 1: Long Papers)}, pages 4582--4597,
  Online. Association for Computational Linguistics.

\bibitem[{Lin et~al.(2022)Lin, Tan, Miller, Tian, and
  Ren}]{lin2022unsupervised}
Bill~Yuchen Lin, Kangmin Tan, Chris Miller, Beiwen Tian, and Xiang Ren. 2022.
\newblock Unsupervised cross-task generalization via retrieval augmentation.
\newblock \emph{arXiv preprint arXiv:2204.07937}.

\bibitem[{Lin(2004)}]{lin-2004-rouge}
Chin-Yew Lin. 2004.
\newblock \href {https://aclanthology.org/W04-1013} {{ROUGE}: A package for
  automatic evaluation of summaries}.
\newblock In \emph{Text Summarization Branches Out}, pages 74--81, Barcelona,
  Spain. Association for Computational Linguistics.

\bibitem[{Liu et~al.(2021)Liu, Shen, Zhang, Dolan, Carin, and
  Chen}]{liu2021makes}
Jiachang Liu, Dinghan Shen, Yizhe Zhang, Bill Dolan, Lawrence Carin, and Weizhu
  Chen. 2021.
\newblock What makes good in-context examples for gpt-3?
\newblock \emph{arXiv preprint arXiv:2101.06804}.

\bibitem[{Liu et~al.(2019)Liu, Ott, Goyal, Du, Joshi, Chen, Levy, Lewis,
  Zettlemoyer, and Stoyanov}]{liu2019roberta}
Yinhan Liu, Myle Ott, Naman Goyal, Jingfei Du, Mandar Joshi, Danqi Chen, Omer
  Levy, Mike Lewis, Luke Zettlemoyer, and Veselin Stoyanov. 2019.
\newblock Roberta: A robustly optimized bert pretraining approach.
\newblock \emph{arXiv preprint arXiv:1907.11692}.

\bibitem[{Lu et~al.(2021)Lu, Bartolo, Moore, Riedel, and
  Stenetorp}]{lu2021fantastically}
Yao Lu, Max Bartolo, Alastair Moore, Sebastian Riedel, and Pontus Stenetorp.
  2021.
\newblock Fantastically ordered prompts and where to find them: Overcoming
  few-shot prompt order sensitivity.
\newblock \emph{arXiv preprint arXiv:2104.08786}.

\bibitem[{McCann et~al.(2018)McCann, Keskar, Xiong, and
  Socher}]{McCann2018decaNLP}
Bryan McCann, Nitish~Shirish Keskar, Caiming Xiong, and Richard Socher. 2018.
\newblock The natural language decathlon: Multitask learning as question
  answering.
\newblock \emph{arXiv preprint arXiv:1806.08730}.

\bibitem[{Mi et~al.(2021)Mi, Zhou, Kong, Cai, Huang, and Faltings}]{mi2021self}
Fei Mi, Wanhao Zhou, Lingjing Kong, Fengyu Cai, Minlie Huang, and Boi Faltings.
  2021.
\newblock \href {https://doi.org/10.18653/v1/2021.emnlp-main.142}
  {Self-training improves pre-training for few-shot learning in task-oriented
  dialog systems}.
\newblock In \emph{Proceedings of the 2021 Conference on Empirical Methods in
  Natural Language Processing}, pages 1887--1898, Online and Punta Cana,
  Dominican Republic. Association for Computational Linguistics.

\bibitem[{Min et~al.(2021)Min, Lewis, Zettlemoyer, and
  Hajishirzi}]{min2021metaicl}
Sewon Min, Mike Lewis, Luke Zettlemoyer, and Hannaneh Hajishirzi. 2021.
\newblock Metaicl: Learning to learn in context.
\newblock \emph{arXiv preprint arXiv:2110.15943}.

\bibitem[{Mishra et~al.(2021)Mishra, Khashabi, Baral, and
  Hajishirzi}]{mishra2021crosstask}
Swaroop Mishra, Daniel Khashabi, Chitta Baral, and Hannaneh Hajishirzi. 2021.
\newblock Cross-task generalization via natural language crowdsourcing
  instructions.
\newblock \emph{arXiv preprint arXiv:2104.08773}.

\bibitem[{Moghe et~al.(2021)Moghe, Steedman, and Birch}]{moghe-etal-2021-cross}
Nikita Moghe, Mark Steedman, and Alexandra Birch. 2021.
\newblock \href {https://doi.org/10.18653/v1/2021.emnlp-main.87} {Cross-lingual
  intermediate fine-tuning improves dialogue state tracking}.
\newblock In \emph{Proceedings of the 2021 Conference on Empirical Methods in
  Natural Language Processing}, pages 1137--1150, Online and Punta Cana,
  Dominican Republic. Association for Computational Linguistics.

\bibitem[{Ott et~al.(2019)Ott, Edunov, Baevski, Fan, Gross, Ng, Grangier, and
  Auli}]{ott-etal-2019-fairseq}
Myle Ott, Sergey Edunov, Alexei Baevski, Angela Fan, Sam Gross, Nathan Ng,
  David Grangier, and Michael Auli. 2019.
\newblock \href {https://doi.org/10.18653/v1/N19-4009} {fairseq: A fast,
  extensible toolkit for sequence modeling}.
\newblock In \emph{Proceedings of the 2019 Conference of the North {A}merican
  Chapter of the Association for Computational Linguistics (Demonstrations)},
  pages 48--53, Minneapolis, Minnesota. Association for Computational
  Linguistics.

\bibitem[{Ouyang et~al.(2022)Ouyang, Wu, Jiang, Almeida, Wainwright, Mishkin,
  Zhang, Agarwal, Slama, Ray et~al.}]{ouyang2022training}
Long Ouyang, Jeff Wu, Xu~Jiang, Diogo Almeida, Carroll~L Wainwright, Pamela
  Mishkin, Chong Zhang, Sandhini Agarwal, Katarina Slama, Alex Ray, et~al.
  2022.
\newblock Training language models to follow instructions with human feedback.
\newblock \emph{Preprint}.

\bibitem[{Paperno et~al.(2016)Paperno, Kruszewski, Lazaridou, Pham, Bernardi,
  Pezzelle, Baroni, Boleda, and Fern{\'a}ndez}]{paperno-etal-2016-lambada}
Denis Paperno, Germ{\'a}n Kruszewski, Angeliki Lazaridou, Ngoc~Quan Pham,
  Raffaella Bernardi, Sandro Pezzelle, Marco Baroni, Gemma Boleda, and Raquel
  Fern{\'a}ndez. 2016.
\newblock \href {https://doi.org/10.18653/v1/P16-1144} {The {LAMBADA} dataset:
  Word prediction requiring a broad discourse context}.
\newblock In \emph{Proceedings of the 54th Annual Meeting of the Association
  for Computational Linguistics (Volume 1: Long Papers)}, pages 1525--1534,
  Berlin, Germany. Association for Computational Linguistics.

\bibitem[{Paszke et~al.(2017)Paszke, Gross, Chintala, Chanan, Yang, DeVito,
  Lin, Desmaison, Antiga, and Lerer}]{paszke2017automatic}
Adam Paszke, Sam Gross, Soumith Chintala, Gregory Chanan, Edward Yang, Zachary
  DeVito, Zeming Lin, Alban Desmaison, Luca Antiga, and Adam Lerer. 2017.
\newblock Automatic differentiation in {PyTorch}.
\newblock In \emph{NIPS Autodiff Workshop}.

\bibitem[{Phang et~al.(2020)Phang, Calixto, Htut, Pruksachatkun, Liu, Vania,
  Kann, and Bowman}]{phang-etal-2020-english}
Jason Phang, Iacer Calixto, Phu~Mon Htut, Yada Pruksachatkun, Haokun Liu, Clara
  Vania, Katharina Kann, and Samuel~R. Bowman. 2020.
\newblock \href {https://aclanthology.org/2020.aacl-main.56} {{E}nglish
  intermediate-task training improves zero-shot cross-lingual transfer too}.
\newblock In \emph{Proceedings of the 1st Conference of the Asia-Pacific
  Chapter of the Association for Computational Linguistics and the 10th
  International Joint Conference on Natural Language Processing}, pages
  557--575, Suzhou, China. Association for Computational Linguistics.

\bibitem[{Phang et~al.(2018)Phang, F{\'e}vry, and Bowman}]{phang2018sentence}
Jason Phang, Thibault F{\'e}vry, and Samuel~R Bowman. 2018.
\newblock Sentence encoders on stilts: Supplementary training on intermediate
  labeled-data tasks.
\newblock \emph{arXiv preprint arXiv:1811.01088}.

\bibitem[{Pilehvar and
  Camacho-Collados(2019)}]{pilehvar-camacho-collados-2019-wic}
Mohammad~Taher Pilehvar and Jose Camacho-Collados. 2019.
\newblock \href {https://doi.org/10.18653/v1/N19-1128} {{W}i{C}: the
  word-in-context dataset for evaluating context-sensitive meaning
  representations}.
\newblock In \emph{Proceedings of the 2019 Conference of the North {A}merican
  Chapter of the Association for Computational Linguistics: Human Language
  Technologies, Volume 1 (Long and Short Papers)}, pages 1267--1273,
  Minneapolis, Minnesota. Association for Computational Linguistics.

\bibitem[{Poth et~al.(2021)Poth, Pfeiffer, R{\"u}ckl{\'e}, and
  Gurevych}]{poth-etal-2021-pre}
Clifton Poth, Jonas Pfeiffer, Andreas R{\"u}ckl{\'e}, and Iryna Gurevych. 2021.
\newblock \href {https://doi.org/10.18653/v1/2021.emnlp-main.827} {{W}hat to
  pre-train on? {E}fficient intermediate task selection}.
\newblock In \emph{Proceedings of the 2021 Conference on Empirical Methods in
  Natural Language Processing}, pages 10585--10605, Online and Punta Cana,
  Dominican Republic. Association for Computational Linguistics.

\bibitem[{Pruksachatkun et~al.(2020)Pruksachatkun, Phang, Liu, Htut, Zhang,
  Pang, Vania, Kann, and Bowman}]{pruksachatkun-etal-2020-intermediate}
Yada Pruksachatkun, Jason Phang, Haokun Liu, Phu~Mon Htut, Xiaoyi Zhang,
  Richard~Yuanzhe Pang, Clara Vania, Katharina Kann, and Samuel~R. Bowman.
  2020.
\newblock \href {https://doi.org/10.18653/v1/2020.acl-main.467}
  {Intermediate-task transfer learning with pretrained language models: When
  and why does it work?}
\newblock In \emph{Proceedings of the 58th Annual Meeting of the Association
  for Computational Linguistics}, pages 5231--5247, Online. Association for
  Computational Linguistics.

\bibitem[{Qin and Eisner(2021)}]{qin-eisner-2021-learning}
Guanghui Qin and Jason Eisner. 2021.
\newblock \href {https://doi.org/10.18653/v1/2021.naacl-main.410} {Learning how
  to ask: Querying {LM}s with mixtures of soft prompts}.
\newblock In \emph{Proceedings of the 2021 Conference of the North American
  Chapter of the Association for Computational Linguistics: Human Language
  Technologies}, pages 5203--5212, Online. Association for Computational
  Linguistics.

\bibitem[{Radford et~al.(2018)Radford, Narasimhan, Salimans, and
  Sutskever}]{radford2018improving}
Alec Radford, Karthik Narasimhan, Tim Salimans, and Ilya Sutskever. 2018.
\newblock Improving language understanding by generative pre-training.

\bibitem[{Raffel et~al.(2020)Raffel, Shazeer, Roberts, Lee, Narang, Matena,
  Zhou, Li, and Liu}]{2020t5}
Colin Raffel, Noam Shazeer, Adam Roberts, Katherine Lee, Sharan Narang, Michael
  Matena, Yanqi Zhou, Wei Li, and Peter~J. Liu. 2020.
\newblock \href {http://jmlr.org/papers/v21/20-074.html} {Exploring the limits
  of transfer learning with a unified text-to-text transformer}.
\newblock \emph{Journal of Machine Learning Research}, 21(140):1--67.

\bibitem[{Reynolds and McDonell(2021)}]{reynolds2021prompt}
Laria Reynolds and Kyle McDonell. 2021.
\newblock Prompt programming for large language models: Beyond the few-shot
  paradigm.
\newblock In \emph{Extended Abstracts of the 2021 CHI Conference on Human
  Factors in Computing Systems}, pages 1--7.

\bibitem[{Roemmele et~al.(2011)Roemmele, Bejan, and
  Gordon}]{roemmele2011choice}
Melissa Roemmele, Cosmin~Adrian Bejan, and Andrew~S. Gordon. 2011.
\newblock Choice of plausible alternatives: An evaluation of commonsense causal
  reasoning.
\newblock In \emph{2011 AAAI Spring Symposium Series}.

\bibitem[{Rubino and Sumita(2020)}]{rubino-sumita-2020-intermediate}
Raphael Rubino and Eiichiro Sumita. 2020.
\newblock \href {https://doi.org/10.18653/v1/2020.coling-main.385}
  {Intermediate self-supervised learning for machine translation quality
  estimation}.
\newblock In \emph{Proceedings of the 28th International Conference on
  Computational Linguistics}, pages 4355--4360, Barcelona, Spain (Online).
  International Committee on Computational Linguistics.

\bibitem[{Sakaguchi et~al.(2020)Sakaguchi, Bras, Bhagavatula, and
  Choi}]{Sakaguchi2020WINOGRANDEAA}
Keisuke Sakaguchi, Ronan~Le Bras, Chandra Bhagavatula, and Yejin Choi. 2020.
\newblock Winogrande: An adversarial winograd schema challenge at scale.
\newblock In \emph{AAAI}.

\bibitem[{Sanh et~al.(2022)Sanh, Webson, Raffel, Bach, Sutawika, Alyafeai,
  Chaffin, Stiegler, Raja, Dey, Bari, Xu, Thakker, Sharma, Szczechla, Kim,
  Chhablani, Nayak, Datta, Chang, Jiang, Wang, Manica, Shen, Yong, Pandey,
  Bawden, Wang, Neeraj, Rozen, Sharma, Santilli, Fevry, Fries, Teehan, Scao,
  Biderman, Gao, Wolf, and Rush}]{sanh2021multitask}
Victor Sanh, Albert Webson, Colin Raffel, Stephen Bach, Lintang Sutawika, Zaid
  Alyafeai, Antoine Chaffin, Arnaud Stiegler, Arun Raja, Manan Dey, M~Saiful
  Bari, Canwen Xu, Urmish Thakker, Shanya~Sharma Sharma, Eliza Szczechla,
  Taewoon Kim, Gunjan Chhablani, Nihal Nayak, Debajyoti Datta, Jonathan Chang,
  Mike Tian-Jian Jiang, Han Wang, Matteo Manica, Sheng Shen, Zheng~Xin Yong,
  Harshit Pandey, Rachel Bawden, Thomas Wang, Trishala Neeraj, Jos Rozen,
  Abheesht Sharma, Andrea Santilli, Thibault Fevry, Jason~Alan Fries, Ryan
  Teehan, Teven~Le Scao, Stella Biderman, Leo Gao, Thomas Wolf, and Alexander~M
  Rush. 2022.
\newblock \href {https://openreview.net/forum?id=9Vrb9D0WI4} {Multitask
  prompted training enables zero-shot task generalization}.
\newblock In \emph{International Conference on Learning Representations}.

\bibitem[{Schick and
  Sch{\"u}tze(2021{\natexlab{a}})}]{schick-schutze-2021-exploiting}
Timo Schick and Hinrich Sch{\"u}tze. 2021{\natexlab{a}}.
\newblock \href {https://aclanthology.org/2021.eacl-main.20} {Exploiting
  cloze-questions for few-shot text classification and natural language
  inference}.
\newblock In \emph{Proceedings of the 16th Conference of the European Chapter
  of the Association for Computational Linguistics: Main Volume}, pages
  255--269, Online. Association for Computational Linguistics.

\bibitem[{Schick and
  Sch{\"u}tze(2021{\natexlab{b}})}]{schick-schutze-2021-shot}
Timo Schick and Hinrich Sch{\"u}tze. 2021{\natexlab{b}}.
\newblock \href {https://doi.org/10.18653/v1/2021.emnlp-main.32} {Few-shot text
  generation with natural language instructions}.
\newblock In \emph{Proceedings of the 2021 Conference on Empirical Methods in
  Natural Language Processing}, pages 390--402, Online and Punta Cana,
  Dominican Republic. Association for Computational Linguistics.

\bibitem[{Schick and
  Sch{\"u}tze(2021{\natexlab{c}})}]{schick-schutze-2021-just}
Timo Schick and Hinrich Sch{\"u}tze. 2021{\natexlab{c}}.
\newblock \href {https://doi.org/10.18653/v1/2021.naacl-main.185} {It{'}s not
  just size that matters: Small language models are also few-shot learners}.
\newblock In \emph{Proceedings of the 2021 Conference of the North American
  Chapter of the Association for Computational Linguistics: Human Language
  Technologies}, pages 2339--2352, Online. Association for Computational
  Linguistics.

\bibitem[{Sorensen et~al.(2022)Sorensen, Robinson, Rytting, Shaw, Rogers,
  Delorey, Khalil, Fulda, and Wingate}]{sorensen2022informationtheoretic}
Taylor Sorensen, Joshua Robinson, Christopher~Michael Rytting, Alexander~Glenn
  Shaw, Kyle~Jeffrey Rogers, Alexia~Pauline Delorey, Mahmoud Khalil, Nancy
  Fulda, and David Wingate. 2022.
\newblock An information-theoretic approach to prompt engineering without
  ground truth labels.
\newblock \emph{arXiv preprint arXiv:2203.11364}.

\bibitem[{Tam et~al.(2021)Tam, R.~Menon, Bansal, Srivastava, and
  Raffel}]{tam2021improving}
Derek Tam, Rakesh R.~Menon, Mohit Bansal, Shashank Srivastava, and Colin
  Raffel. 2021.
\newblock \href {https://doi.org/10.18653/v1/2021.emnlp-main.407} {Improving
  and simplifying pattern exploiting training}.
\newblock In \emph{Proceedings of the 2021 Conference on Empirical Methods in
  Natural Language Processing}, pages 4980--4991, Online and Punta Cana,
  Dominican Republic. Association for Computational Linguistics.

\bibitem[{Vu et~al.(2021)Vu, Luong, Le, Simon, and Iyyer}]{vu-etal-2021-strata}
Tu~Vu, Minh-Thang Luong, Quoc Le, Grady Simon, and Mohit Iyyer. 2021.
\newblock \href {https://doi.org/10.18653/v1/2021.emnlp-main.462} {{ST}ra{TA}:
  Self-training with task augmentation for better few-shot learning}.
\newblock In \emph{Proceedings of the 2021 Conference on Empirical Methods in
  Natural Language Processing}, pages 5715--5731, Online and Punta Cana,
  Dominican Republic. Association for Computational Linguistics.

\bibitem[{Wang et~al.(2019)Wang, Pruksachatkun, Nangia, Singh, Michael, Hill,
  Levy, and Bowman}]{superglue}
Alex Wang, Yada Pruksachatkun, Nikita Nangia, Amanpreet Singh, Julian Michael,
  Felix Hill, Omer Levy, and Samuel Bowman. 2019.
\newblock \href
  {https://proceedings.neurips.cc/paper/2019/file/4496bf24afe7fab6f046bf4923da8de6-Paper.pdf}
  {Superglue: A stickier benchmark for general-purpose language understanding
  systems}.
\newblock In \emph{Advances in Neural Information Processing Systems},
  volume~32. Curran Associates, Inc.

\bibitem[{Wang et~al.(2021{\natexlab{a}})Wang, Fang, Khabsa, Mao, and
  Ma}]{wang2021entailment}
Sinong Wang, Han Fang, Madian Khabsa, Hanzi Mao, and Hao Ma.
  2021{\natexlab{a}}.
\newblock Entailment as few-shot learner.
\newblock \emph{arXiv preprint arXiv:2104.14690}.

\bibitem[{Wang et~al.(2021{\natexlab{b}})Wang, Mukherjee, Liu, Gao, Awadallah,
  and Gao}]{wang2021list}
Yaqing Wang, Subhabrata Mukherjee, Xiaodong Liu, Jing Gao, Ahmed~Hassan
  Awadallah, and Jianfeng Gao. 2021{\natexlab{b}}.
\newblock List: Lite self-training makes efficient few-shot learners.
\newblock \emph{arXiv preprint arXiv:2110.06274}.

\bibitem[{Wei et~al.(2022)Wei, Bosma, Zhao, Guu, Yu, Lester, Du, Dai, and
  Le}]{wei2021finetuned}
Jason Wei, Maarten Bosma, Vincent Zhao, Kelvin Guu, Adams~Wei Yu, Brian Lester,
  Nan Du, Andrew~M. Dai, and Quoc~V Le. 2022.
\newblock \href {https://openreview.net/forum?id=gEZrGCozdqR} {Finetuned
  language models are zero-shot learners}.
\newblock In \emph{International Conference on Learning Representations}.

\bibitem[{Weller et~al.(2020)Weller, Lourie, Gardner, and
  Peters}]{weller-etal-2020-learning}
Orion Weller, Nicholas Lourie, Matt Gardner, and Matthew~E. Peters. 2020.
\newblock \href {https://doi.org/10.18653/v1/2020.emnlp-main.105} {Learning
  from task descriptions}.
\newblock In \emph{Proceedings of the 2020 Conference on Empirical Methods in
  Natural Language Processing (EMNLP)}, pages 1361--1375, Online. Association
  for Computational Linguistics.

\bibitem[{Xu et~al.(2022)Xu, Chen, Du, Shao, Wang, Li, and
  Yang}]{xu2022zeroprompt}
Hanwei Xu, Yujun Chen, Yulun Du, Nan Shao, Yanggang Wang, Haiyu Li, and Zhilin
  Yang. 2022.
\newblock Zeroprompt: Scaling prompt-based pretraining to 1,000 tasks improves
  zero-shot generalization.
\newblock \emph{arXiv preprint arXiv:2201.06910}.

\bibitem[{Ye et~al.(2021)Ye, Lin, and Ren}]{ye2021crossfit}
Qinyuan Ye, Bill~Yuchen Lin, and Xiang Ren. 2021.
\newblock \href {https://doi.org/10.18653/v1/2021.emnlp-main.572}
  {{C}ross{F}it: A few-shot learning challenge for cross-task generalization in
  {NLP}}.
\newblock In \emph{Proceedings of the 2021 Conference on Empirical Methods in
  Natural Language Processing}, pages 7163--7189, Online and Punta Cana,
  Dominican Republic. Association for Computational Linguistics.

\bibitem[{Yin et~al.(2019)Yin, Hay, and Roth}]{yin-etal-2019-benchmarking}
Wenpeng Yin, Jamaal Hay, and Dan Roth. 2019.
\newblock \href {https://doi.org/10.18653/v1/D19-1404} {Benchmarking zero-shot
  text classification: Datasets, evaluation and entailment approach}.
\newblock In \emph{Proceedings of the 2019 Conference on Empirical Methods in
  Natural Language Processing and the 9th International Joint Conference on
  Natural Language Processing (EMNLP-IJCNLP)}, pages 3914--3923, Hong Kong,
  China. Association for Computational Linguistics.

\bibitem[{Yin et~al.(2020)Yin, Rajani, Radev, Socher, and
  Xiong}]{yin-etal-2020-universal}
Wenpeng Yin, Nazneen~Fatema Rajani, Dragomir Radev, Richard Socher, and Caiming
  Xiong. 2020.
\newblock \href {https://doi.org/10.18653/v1/2020.emnlp-main.660} {Universal
  natural language processing with limited annotations: Try few-shot textual
  entailment as a start}.
\newblock In \emph{Proceedings of the 2020 Conference on Empirical Methods in
  Natural Language Processing (EMNLP)}, pages 8229--8239, Online. Association
  for Computational Linguistics.

\bibitem[{Zhang et~al.(2021)Zhang, Bui, Yoon, Chen, Liu, Xia, Tran, Chang, and
  Yu}]{zhang2021few}
Jianguo Zhang, Trung Bui, Seunghyun Yoon, Xiang Chen, Zhiwei Liu, Congying Xia,
  Quan~Hung Tran, Walter Chang, and Philip Yu. 2021.
\newblock \href {https://doi.org/10.18653/v1/2021.emnlp-main.144} {Few-shot
  intent detection via contrastive pre-training and fine-tuning}.
\newblock In \emph{Proceedings of the 2021 Conference on Empirical Methods in
  Natural Language Processing}, pages 1906--1912, Online and Punta Cana,
  Dominican Republic. Association for Computational Linguistics.

\bibitem[{Zhang et~al.(2020)Zhang, Zhao, Saleh, and Liu}]{pmlr-v119-zhang20ae}
Jingqing Zhang, Yao Zhao, Mohammad Saleh, and Peter Liu. 2020.
\newblock \href {https://proceedings.mlr.press/v119/zhang20ae.html} {{PEGASUS}:
  Pre-training with extracted gap-sentences for abstractive summarization}.
\newblock In \emph{Proceedings of the 37th International Conference on Machine
  Learning}, volume 119 of \emph{Proceedings of Machine Learning Research},
  pages 11328--11339. PMLR.

\bibitem[{Zhang et~al.(2022)Zhang, Li, Chen, Deng, Bi, Tan, Huang, and
  Chen}]{zhang2022differentiable}
Ningyu Zhang, Luoqiu Li, Xiang Chen, Shumin Deng, Zhen Bi, Chuanqi Tan, Fei
  Huang, and Huajun Chen. 2022.
\newblock \href {https://openreview.net/forum?id=ek9a0qIafW} {Differentiable
  prompt makes pre-trained language models better few-shot learners}.
\newblock In \emph{International Conference on Learning Representations}.

\bibitem[{Zhang et~al.(2018)Zhang, Liu, Liu, Gao, Duh, and
  Durme}]{zhang2018record}
Sheng Zhang, Xiaodong Liu, Jingjing Liu, Jianfeng Gao, Kevin Duh, and
  Benjamin~Van Durme. 2018.
\newblock {ReCoRD}: Bridging the gap between human and machine commonsense
  reading comprehension.
\newblock \emph{arXiv preprint 1810.12885}.

\bibitem[{Zhang and Zhang(2021)}]{zhang-zhang-2021-qa}
Shiwei Zhang and Xiuzhen Zhang. 2021.
\newblock \href {https://aclanthology.org/2021.alta-1.16} {Does {QA}-based
  intermediate training help fine-tuning language models for text
  classification?}
\newblock In \emph{Proceedings of the The 19th Annual Workshop of the
  Australasian Language Technology Association}, pages 158--162, Online.
  Australasian Language Technology Association.

\bibitem[{Zhao et~al.(2021)Zhao, Wallace, Feng, Klein, and
  Singh}]{pmlr-v139-zhao21c}
Zihao Zhao, Eric Wallace, Shi Feng, Dan Klein, and Sameer Singh. 2021.
\newblock \href {https://proceedings.mlr.press/v139/zhao21c.html} {Calibrate
  before use: Improving few-shot performance of language models}.
\newblock In \emph{Proceedings of the 38th International Conference on Machine
  Learning}, volume 139 of \emph{Proceedings of Machine Learning Research},
  pages 12697--12706. PMLR.

\bibitem[{Zhong et~al.(2021)Zhong, Lee, Zhang, and Klein}]{leeadapting}
Ruiqi Zhong, Kristy Lee, Zheng Zhang, and Dan Klein. 2021.
\newblock \href {https://doi.org/10.18653/v1/2021.findings-emnlp.244} {Adapting
  language models for zero-shot learning by meta-tuning on dataset and prompt
  collections}.
\newblock In \emph{Findings of the Association for Computational Linguistics:
  EMNLP 2021}, pages 2856--2878, Punta Cana, Dominican Republic. Association
  for Computational Linguistics.

\bibitem[{Zhou et~al.(2019)Zhou, Khashabi, Ning, and
  Roth}]{zhou-etal-2019-going}
Ben Zhou, Daniel Khashabi, Qiang Ning, and Dan Roth. 2019.
\newblock \href {https://doi.org/10.18653/v1/D19-1332} {{``}going on a
  vacation{''} takes longer than {``}going for a walk{''}: A study of temporal
  commonsense understanding}.
\newblock In \emph{Proceedings of the 2019 Conference on Empirical Methods in
  Natural Language Processing and the 9th International Joint Conference on
  Natural Language Processing (EMNLP-IJCNLP)}, pages 3363--3369, Hong Kong,
  China. Association for Computational Linguistics.

\end{thebibliography}
